\documentclass{article}

\usepackage{arxiv}
\usepackage[utf8]{inputenc}
\usepackage[T1]{fontenc}
\usepackage{hyperref}
\usepackage{url}
\usepackage{booktabs}
\usepackage{amsfonts}
\usepackage{amsmath}
\usepackage{amssymb}
\usepackage{nicefrac}
\usepackage{microtype}
\usepackage{graphicx}
\usepackage{float}
\usepackage{subcaption}
\usepackage{multirow}
\usepackage{pifont}
\usepackage[capitalize,noabbrev]{cleveref}
\usepackage{titlesec}

\hypersetup{colorlinks=true, urlcolor=blue, linkcolor=black, citecolor=black}

\graphicspath{ {./} }

\newcommand{\eg}{\emph{e.g.}}

\newcommand{\etal}{\emph{et al.}}

\newcommand{\cmark}{\ding{51}}
\newcommand{\xmark}{\ding{55}}

\title{EgoTraj: Real-World Egocentric Human Trajectory Dataset for Multimodal Prediction}

\author{%
  \textbf{Ahmad Yehia}\textsuperscript{1} \quad
  \textbf{Abduallah Mohamed}\textsuperscript{2}\thanks{Work done outside of Meta in a personal capacity.} \quad
  \textbf{Tianyi Wang}\textsuperscript{1} \quad
  \textbf{Jiseop Byeon}\textsuperscript{1} \quad
  \textbf{Kun Qian}\textsuperscript{1} \\[3pt]
  \textbf{Junfeng Jiao}\textsuperscript{3} \quad
  \textbf{Christian Claudel}\textsuperscript{1,}\thanks{Corresponding author.}
  \\[6pt]
  \normalfont\small
  \textsuperscript{1}Department of Civil, Architectural, and Environmental Engineering,
  The University of Texas at Austin, Austin, TX 78712, USA \\
  \textsuperscript{2}Meta Reality Labs \\
  \textsuperscript{3}School of Architecture, The University of Texas at Austin, Austin, TX 78712, USA \\[3pt]
  \texttt{\{ahmad.yehia, bonny.wang, jsbyeon, kunqian, christian.claudel\}@utexas.edu} \\
  \texttt{jjiao@austin.utexas.edu} \quad \texttt{abduallahadel@meta.com}
}

\begin{document}
\maketitle


\begin{abstract}
Accurately forecasting human trajectories from an egocentric perspective plays a central role in applications such as humanoid robotics, wearable sensing systems, and assistive navigation.
However, progress in this direction remains limited due to the scarcity of egocentric trajectory datasets collected in real-world environments.
Addressing this need, we introduce EgoTraj, an egocentric multimodal open dataset recorded using Meta Quest Pro (MQPro).
EgoTraj contains 75 sequences of human navigation collected from multiple MQPro wearers in real-world urban environments.
Each recording provides synchronized RGB video along with ground-truth data, including continuous time-synchronized 6-degree-of-freedom head poses, per-frame 3D eye gaze vectors, scene annotations.
To the best of our knowledge, EgoTraj differs from typical egocentric trajectory datasets by capturing long-horizon, self-directed navigation across diverse urban routes with broad participant diversity.
To demonstrate the potential of the dataset, we benchmark several state-of-the-art methods for egocentric trajectory prediction and conduct ablation studies to analyze the contributions of gaze, scene, and motion cues. The results highlight the utility of EgoTraj for AR-based perception, navigation, and assistive systems. The EgoTraj dataset, code, and \textit{EgoViz Dashboard} are publicly available at \url{https://github.com/yehiahmad/EgoTraj}.

\end{abstract}


\begin{figure}[t]
  \centering
  \includegraphics[width=\linewidth]{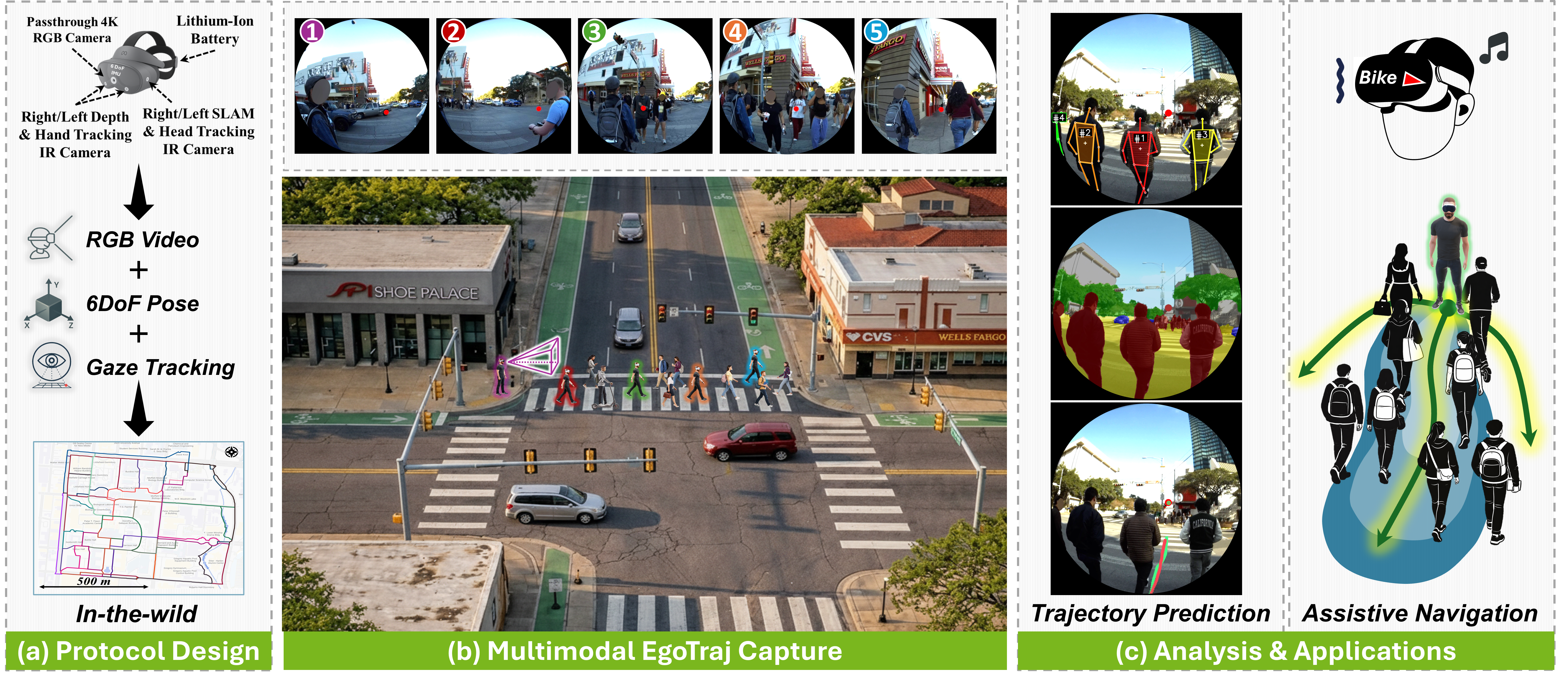}
    \caption{Overview of EgoTraj.
    (a) \textbf{Protocol design:} the Meta Quest Pro headset records synchronized RGB video, 6DoF head pose, and gaze signals during in-the-wild navigation.
    (b) \textbf{Multimodal EgoTraj capture:} representative egocentric frames from a crosswalk navigation scenario with multiple social interactions.
    (c) \textbf{Analysis and applications:} the dataset supports downstream tasks such as assistive navigation for blind and visually impaired users.}
  \label{fig:intro}
\end{figure}

\section{Introduction}
\label{sec:intro}

Egocentric perception forms the foundation of human navigation behavior \cite{yehia2025arcas}.
Humans naturally perceive and interpret their surroundings from an egocentric viewpoint to guide decisions about subsequent motions \cite{chen2019crowd}.
Replicating this effortless human capability in machines, however, remains a long-standing challenge in embodied AI and computer vision \cite{marchetti2020multiple}.
Learning such perceptual cues plays a pivotal role in forecasting the trajectory of humanoid robots~\cite{peng2025real,kim2024armor,pan2025lookout}, AR-VR systems~\cite{escobar2025egocast}, and autonomous vehicles~\cite{zheng2024genad,hu2023planning,jain2020discrete}.
At the same time, recent advances in egocentric vision and wearable sensing technologies have facilitated the development of assistive wayfinding systems, including blind navigation~\cite{tian2013wearable}, object echo-location~\cite{tang2014assistive}, and personalized object recognition~\cite{kacorri2017people}.

As augmented reality (AR) technologies evolve, assistive systems increasingly leverage egocentric observations to interpret the user's surroundings and provide context-aware guidance for subsequent actions.
For instance, by modeling navigation behaviors learned from sighted individuals, forecasting the short-term trajectory can help blind or visually impaired users navigate more independently and safely in social spaces~\cite{qiu2022egocentric}.
Moreover, using synchronized multimodal sensors, predicted trajectories can further be conveyed to the headset users through visual overlays~\cite{yehia2025arcas}, haptic feedback~\cite{wang2017enabling}, or auditory instructions~\cite{google2021guideline}.
Beyond assistive applications, insights from human navigation behavior provide valuable guidance for social-aware robotic navigation.
Humans naturally adhere to implicit social conventions when navigating crowded environments, such as maintaining interpersonal distance, yielding in constrained spaces, and adjusting motion based on gaze and attention cues.
By modeling these behaviors, mobile robots~\cite{wang2025xr} and humanoid robots~\cite{qiu2022egocentric} can learn navigation strategies that are not only collision-free but also socially compliant, enabling safer operation in human-centered environments.

Existing human trajectory prediction methods have made remarkable progress; however, this progress largely relies on datasets captured from bird's-eye viewpoints (BEV) or static cameras \cite{alahi2016social,mohamed2022social}.
Such datasets capture only externally observable motion and fail to model how
humans perceive, plan, and initiate their motion from a first-person perspective.
Consequently, most trajectory forecasting methods~\cite{alahi2016social,shi2023trajectory} rely primarily on motion history and scene context rather than intention-related cues.
In contrast, egocentric vision provides direct access to how humans perceive their
surroundings and navigate through them.
Over the last two decades, egocentric vision research has produced large-scale datasets spanning diverse activities and environments~\cite{grauman2022ego4d,pan2023aria,lv2024aria,grauman2024ego}.
However, these datasets are primarily designed for action recognition and video understanding, with limited emphasis on forecasting the wearer's trajectory for navigation.
As a result, several works have introduced egocentric datasets for trajectory prediction~\cite{pan2025lookout,qiu2025egocognav}.
Nevertheless, many of these egocentric trajectory datasets remain limited in scale, gathered from a single participant, restricted to indoor environments, or lacking synchronized gaze data necessary for intention-aware trajectory forecasting.
Critically, only a few leverage AR-style sensing devices, which introduce realistic challenges such as rapid ego-motion, viewpoint changes, unstable visual observations, and low-power sensing hardware.

Motivated by the lack of real-world egocentric trajectory datasets, we introduce EgoTraj, a multimodal dataset designed to advance research in egocentric trajectory forecasting and assistive AR navigation. To the best of our knowledge, EgoTraj is the first large-scale egocentric trajectory dataset to jointly provide synchronized 6DoF head pose, per-frame 3D gaze vectors, egocentric RGB video, and scene annotations, collected from various participants navigating self-chosen routes across pedestrian crossings, crowded sidewalks, and busy streets.

The main contributions of this work are summarized as follows:
\begin{itemize}
\item \textbf{EgoTraj Dataset.} We introduce a large-scale multimodal dataset for egocentric trajectory forecasting collected by Meta Quest Pro (MQPro) wearers navigating in real-world traffic environments, containing synchronized 6DoF head pose, gaze tracking data, egocentric RGB video, and scene annotations.
\item \textbf{Benchmark and Evaluation Protocol.} We establish a standardized benchmark for egocentric trajectory forecasting by evaluating several state-of-the-art prediction models for both qualitative and quantitative analyses.
\item \textbf{Multimodal Analysis.} Through ablation studies, we analyze the impact of gaze, social interactions, and scene context on trajectory prediction under realistic egocentric conditions.
\end{itemize}

\section{Related Work}
\label{sec:related}

\subsection{Non-Egocentric Human Trajectory Datasets}

Most human trajectory prediction studies rely on a third-person perspective, where scenes are captured either from a BEV or stationary cameras.
In particular, the ETH~\cite{pellegrini2009you} and UCY~\cite{lerner2007crowds} datasets have become the dominant benchmarks for trajectory prediction, with a total of 1,536 tracked pedestrians covering trajectory patterns such as people walking together, crossing paths, and forming or separating into groups.
However, these benchmarks remain limited in scene diversity and restricted to pedestrian-only trajectories.
To address these limitations, the Stanford Drone Dataset~\cite{robicquet2017learning} and inD~\cite{bock2020ind} expanded dataset scale using drone footage to capture the movements of multiple road users, including pedestrians, bicyclists, skateboarders, and vehicles.
In parallel, the autonomous driving domain introduced large-scale ego-vehicle datasets such as nuScenes~\cite{caesar2020nuscenes}, Waymo Open~\cite{ettinger2021large}, and JRDB~\cite{martin2021jrdb}.
Despite their scale and diversity, these datasets remain fundamentally third-person in nature.
While they accurately capture where pedestrians move, they do not model how individuals perceive the scene or how gaze and visual attention shape future actions.
This limitation motivates the development of egocentric trajectory datasets that integrate motion with perceptual cues, such as gaze and head pose, for intent-aware AR applications.

\begin{table}[t]
\centering
\caption{Comparison of egocentric trajectory datasets across recording settings, scale, and sensing modalities.
$^{\dagger}$\,Uses GPS-only localization.
$^{\ddagger}$\,Reconstructs trajectories offline using structure-from-motion.
$^{\ast}$\,Captures activity-level trajectories rather than pedestrian navigation.}
\label{tab:ego-comparison}
\begingroup
\setlength{\tabcolsep}{4.2pt}
\renewcommand{\arraystretch}{1.15}
\resizebox{\textwidth}{!}{%
\begin{tabular}{lccccccccc}
\toprule
 & & & & & & & \multicolumn{2}{c}{\textbf{Modalities}} & \\
\cmidrule(lr){8-9}
\textbf{Dataset} & \textbf{Year} & \textbf{Setting} & \textbf{Hours} & \textbf{Frames} & \textbf{Subj.} &
\textbf{Device} & \textbf{Gaze} & \textbf{6DoF} & \textbf{Scene Ann.} \\
\midrule
KrishnaCam$^\dagger$~\cite{singh2016krishnacam} & 2016 & Outdoor & 70.0 & 7.6M & 1 &
Google Glass & \xmark & \xmark & \xmark \\
EgoMotion$^\ddagger$~\cite{park2016egocentric} & 2016 & In+Out & 9.1 & 65.5K & N/P &
GoPro Stereo & \xmark & \xmark & \xmark \\
FPL$^\ast$~\cite{yagi2018future} & 2018 & Outdoor & 4.5 & 162K & N/P &
Chest Cam & \xmark & \xmark & \xmark \\
Nymeria~\cite{ma2024nymeria} & 2024 & In+Out & 300 & 32.4M & 264 &
Aria + Xsens & \cmark & \cmark & \cmark \\
EgoNav~\cite{wang2024egonav} & 2024 & In+Out & 3.3 & 237.6K & N/P &
RealSense & \xmark & \cmark & \xmark \\
LookOut~\cite{pan2025lookout} & 2025 & In+Out & 4.0 & 288K & N/P &
Aria & \cmark & \cmark & \xmark \\
EgoCogNav~\cite{qiu2025egocognav} & 2025 & In+Out & 6.0 & 432K & 17 &
Tobii + Aria & \cmark & \cmark & \xmark \\
\midrule
\textbf{EgoTraj (ours)} & \textbf{2026} & \textbf{Outdoor} & \textbf{10.7} & \textbf{1.15M} & \textbf{75} &
\textbf{Quest Pro} & \cmark & \cmark & \cmark \\
\bottomrule
\end{tabular}
}%
\endgroup
\end{table}

\subsection{Egocentric Human Trajectory Datasets}
Recent advances in wearable sensors have enabled data collection from a first-person perspective, accelerating research in egocentric vision~\cite{pan2023aria,lv2024aria}.
However, a limited subset of egocentric datasets explicitly addresses pedestrian navigation.
KrishnaCam~\cite{singh2016krishnacam} provides egocentric video streams with GPS position, acceleration, and body orientation collected by one participant, while Yagi~\etal~\cite{yagi2018future} record first-person videos from a chest-mounted camera to predict future 2D locations of nearby pedestrians.
However, neither dataset included gaze or ground-truth 6DoF pose of the camera wearer, limiting their behavioral diversity and applications for gaze-informed trajectory prediction.
Park~\etal's EgoMotion dataset~\cite{park2016egocentric} pioneers egocentric future localization using GoPro stereo cameras across 26 indoor and outdoor scenes.
However, trajectories are reconstructed offline via structure from motion rather than tracked in real time, and the dataset includes neither gaze nor pose telemetry.
These limitations have motivated subsequent efforts that incorporate AR glasses and additional sensing modalities.

Project Aria datasets such as the Aria Digital Twin (ADT)~\cite{pan2023aria}, Aria Everyday Activities (AEA)~\cite{lv2024aria}, and Nymeria~\cite{ma2024nymeria} paired egocentric video with 6DoF pose and gaze across various activities.
However, ADT was restricted to indoor activities, AEA focused on single-human scenarios, and Nymeria captured multiple actors only in collaborative settings.
LookOut~\cite{pan2025lookout} introduced a 4-hour navigation dataset with 6D head pose trajectories in real-world pedestrian scenarios, yet did not record gaze.
The dataset most closely related to our work is the Cognition-Aware Egocentric Navigation (CEN) dataset from EgoCogNav~\cite{qiu2025egocognav}, which paired gaze with walking trajectories across 6~hours of real-world navigation.
However, CEN is designed for cognition-aware modeling and perceived navigational uncertainty.
In addition, the 6-hour recordings are not exclusive to outdoor environments.
EgoTraj addresses a key gap by providing synchronized 6DoF pose, gaze tracking, and egocentric RGB video augmented with scene annotations, collected from 75~participants navigating self-chosen routes in real urban environments.
EgoTraj enables large-scale study of gaze-informed egocentric trajectory prediction in naturalistic pedestrian environments. The detailed comparison of the existing egocentric trajectory datasets and the proposed EgoTraj dataset is shown in Table~\ref{tab:ego-comparison}.

\section{Dataset Description}
\label{sec:dataset}

\subsection{Dataset Collection}

\paragraph{Hardware.}
Each participant wears an MQPro headset operating in full-color passthrough mode.
MQPro offers a non-intrusive, cost-effective, and readily deployable alternative to prior works that rely on custom-built sensor suites~\cite{nguyen2023toward,wang2024egonav} or teleoperated robotic platforms~\cite{karnan2022socially}.
The headset integrates a passthrough RGB camera, two infrared eye-tracking cameras, four inside-out tracking cameras, and a 6-axis IMU, as illustrated in \cref{fig:intro} (a).
A custom Unity application interfaces with the built-in visual--inertial SLAM system and records time-synchronized data at 30\,Hz, including 6DoF head pose, 3D gaze origin and direction vectors, and egocentric RGB video.
During each session, participants use an MQPro controller to start and terminate data recording.

\paragraph{Recording Protocol.}
Before each session, participants completed a brief orientation covering the study objectives, experimental procedure, headset fitting, and eye-tracking calibration.
During data collection, participants navigated between two of seven predefined outdoor waypoints distributed across urban areas, including sidewalks, crosswalks, and busy streets (\cref{fig:intro} (a)).
In contrast to fully scripted protocols, participants were free to choose their own routes between waypoints, enabling naturalistic route-finding behavior, spontaneous interactions with surrounding traffic conditions and realistic decision-making during navigation.
Each session was capped at 15\,minutes (8\,min on average) to promote diverse navigation patterns across participants.
Participants were instructed to obey traffic rules, check for passing vehicles and traffic signals before crossing streets, to avoid indoor shortcuts, and terminate the session immediately if they experienced any discomfort.
A researcher followed each participant at a safe distance to monitor the session and verify waypoint completion without influencing navigation behavior.
Importantly, the seven waypoints denote landmarks rather than pinned GPS points, yielding 21 origin--destination pairs, and 31 of the 75 participants were unfamiliar with the recording area.
To verify that participants did not follow templated routes between the same landmarks, we computed pairwise Dynamic Time Warping (DTW) distances within each of the 21 waypoint pairs ($n=103$ within-pair comparisons), obtaining a median of $122.8$\,m with a range of $[19.3,\,379.4]$\,m.
This wide spread indicates that walkers chose meaningfully different paths between identical endpoint pairs rather than retracing a scripted route.

\begin{figure}[t]
  \centering
  \includegraphics[width=\linewidth]{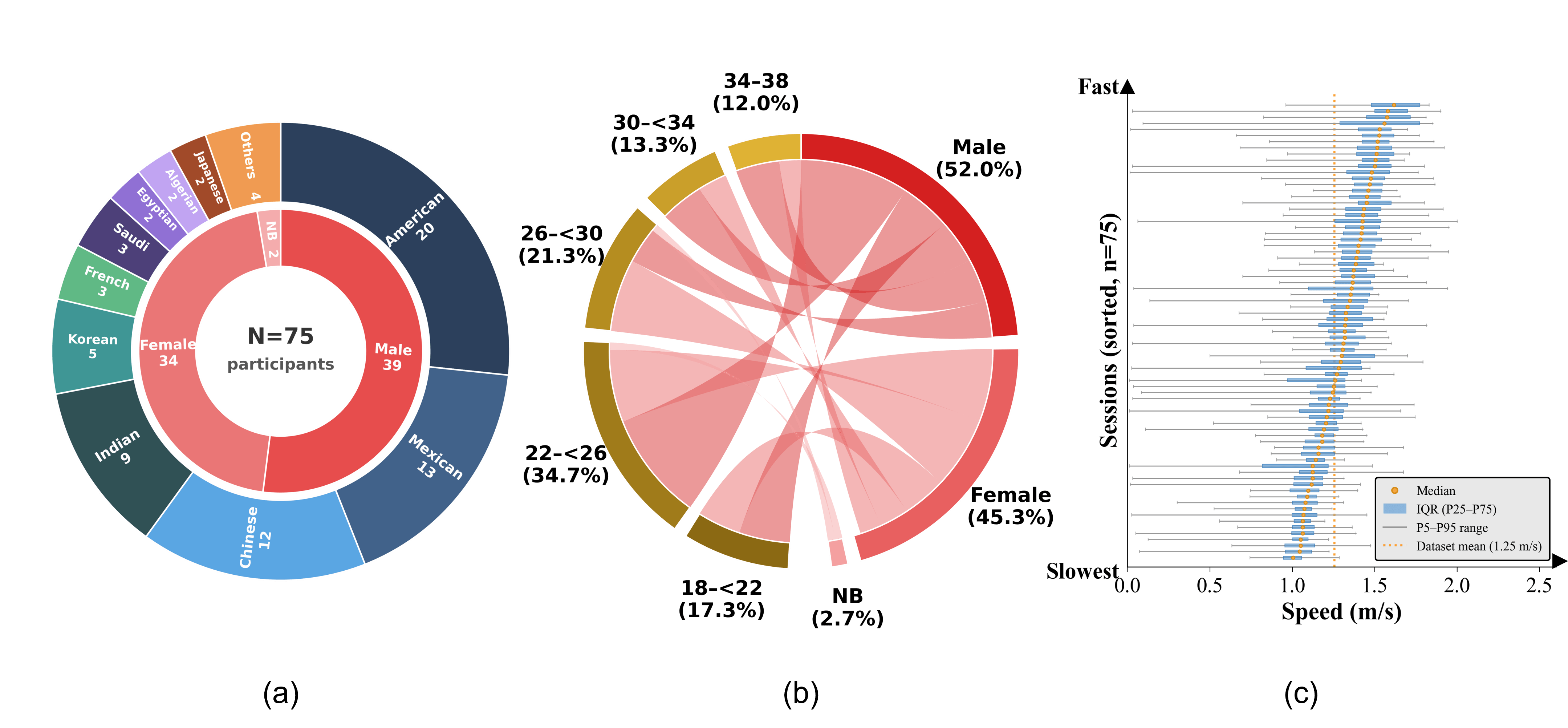}
    \caption{Dataset statistics of EgoTraj participants.
    (a) Nationality breakdown of the 75 recruited participants.
    (b) Age and gender composition across participants.
    (c) Walking speed distribution observed during navigation sessions.}
    \label{fig:data_stat}
\end{figure}

\paragraph{Participants.}
We recruited $N=75$ volunteers to capture diverse navigation behaviors within the dataset.
Eligibility criteria required adults with normal or corrected-to-normal vision and the ability to walk outdoors unassisted.
Each participant contributed exactly one recording session, and no individual appears more than once in the dataset.
Participants received monetary compensation upon completion.
The cohort was balanced across gender to reduce demographic bias, with participants ranging from 18 to 38 years old and representing 14 nationalities (\cref{fig:data_stat} (b)).

\paragraph{Privacy and Ethics.}
This study was approved by an Institutional Review Board (IRB).
All volunteers provided written informed consent prior to data collection and retained the right to withdraw from the study, request deletion of their data, or redact specific portions of their recordings.
To satisfy de-identification requirements for personally identifiable information (PII), all recordings were processed using EgoBlur~\cite{raina2023aria} to automatically detect and blur faces and vehicle license plates.
Session identifiers consist of anonymized date--time codes with no linkage to personal information, and raw unprocessed videos are not publicly distributed.

\subsection{Data Processing}
To prepare the public release, we implemented a multi-stage processing pipeline.
Raw telemetry from the MQPro is recorded as a single CSV file per session at 30\,Hz, containing 6DoF head pose (position, orientation quaternion, linear velocity, and angular velocity) together with binocular gaze signals (3D origin and direction vectors).
Egocentric RGB video is simultaneously recorded as H.264-encoded MP4 segments at 30\,fps.
Because certain sensor streams (\eg, IMU) operate at higher frequencies than the RGB camera, all telemetry signals are resampled to align with the 30\,Hz video timeline.
Synchronized ground truth is obtained by matching each RGB frame to the closest telemetry timestamp.
The finalized dataset is packaged as a single HDF5 file per session with three groups: \textit{pose} (timestamp, position, rotation, linear velocity, angular velocity), \textit{gaze} (origin, direction), and \textit{video} (segment index, frame index).
This processed format enables efficient frame-level indexing and direct retrieval of each RGB frame with its corresponding motion and gaze data.
The processed HDF5 files with privacy-blurred RGB frames are included in the public release.

\subsection{Dataset Statistics}
EgoTraj comprises 75 recording sessions totaling 10.7\, accumulated recording hours of synchronized data with 1.15M RGB frames (\cref{tab:ego-comparison}).
Across all sessions, participants walked a cumulative distance of 46.73\,km in diverse urban environments, with individual session durations ranging from 5 to 15\,minutes.
Recordings were collected during both moderate and high pedestrian traffic conditions.
\Cref{fig:data_stat} (c) shows the per-session speed distribution sorted by median speed.
The dataset-wide mean walking speed is 1.25\,m/s, consistent with typical pedestrian walking speeds (1.2--1.4\,m/s).
Lower-speed segments generally correspond to congested areas such as busy intersections or narrow walkways, while higher-speed sessions occur along straight sidewalks or during hurry-up segments before traffic signals change.
Across the dataset, 7\% of frames exhibit near-stationary motion ($<0.3$\,m/s), capturing stop-and-go pedestrian dynamics.

\subsection{Scene Annotation}
\label{sec:vlm}
\begin{figure}[t]
  \centering
  \includegraphics[width=\linewidth]{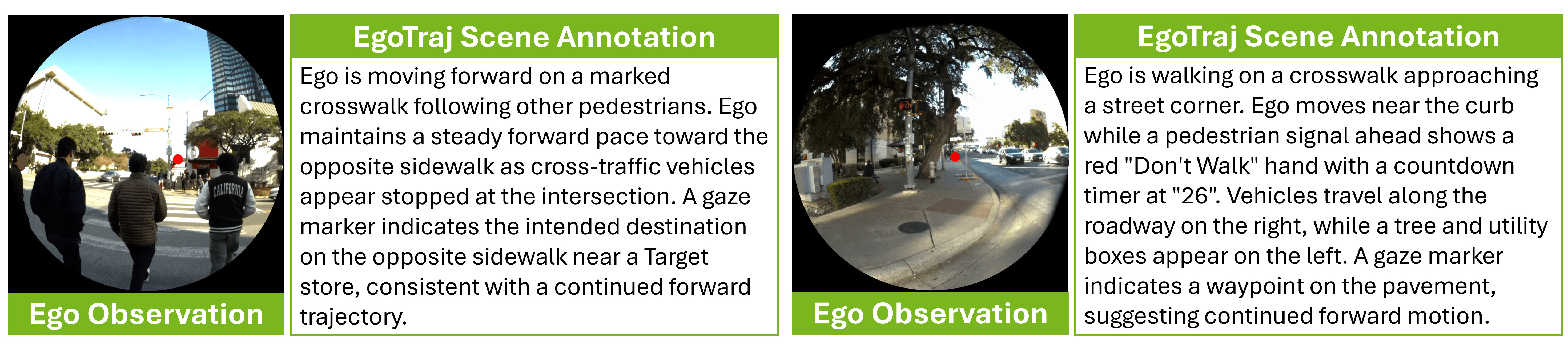}
  \caption{Examples of VLM-generated scene annotations. Each tab shows an egocentric frame with the gaze marker (red dot) alongside the structured annotation produced by Qwen2.5-VL-7B.}
  \label{fig:annotation}
\end{figure}

To augment EgoTraj with high-level contextual cues, we generated structured egocentric scene descriptions using a vision-language model (VLM) pipeline.
The annotations target navigation-relevant elements in urban walking environments, including surrounding context, traffic activity, gaze fixation targets, and inferred short-term movement intent.
These language descriptions complement the trajectory data by capturing behavioral context, such as following pedestrian flow or approaching an intersection, while remaining consistent with the underlying multimodal signals.
From 253 privacy-blurred video segments, frames were sampled at 1\,fps, yielding 38,606 annotated frames.
Each frame was processed using Qwen2.5-VL-7B-Instruct~\cite{wang2024qwen2}.
An example-driven prompt enforces a consistent egocentric narration style that prioritizes interaction and navigation cues over generic scene descriptions (\cref{fig:annotation}). Annotation reliability was evaluated in two stages.
First, structural compliance was measured on 100 stratified frames sampled across sessions.
Compliance required adherence to the prescribed tag structure and inclusion of an explicit navigation decision based on corresponding motion data.
Using chain-of-thoughts (CoT), after up to two prompt retries for non-conforming outputs, the pipeline achieved 96\% compliance.
Second, these annotations were independently reviewed by two human annotators to verify semantic consistency with the corresponding video frame, achieving 93\% inter-annotator agreement.
Final annotations were stored as JSON sidecar files indexed by session and frame identifier alongside the HDF5 dataset.

\subsection{Gaze Calibration}
\begin{figure}[t]
  \centering
  \includegraphics[width=\linewidth]{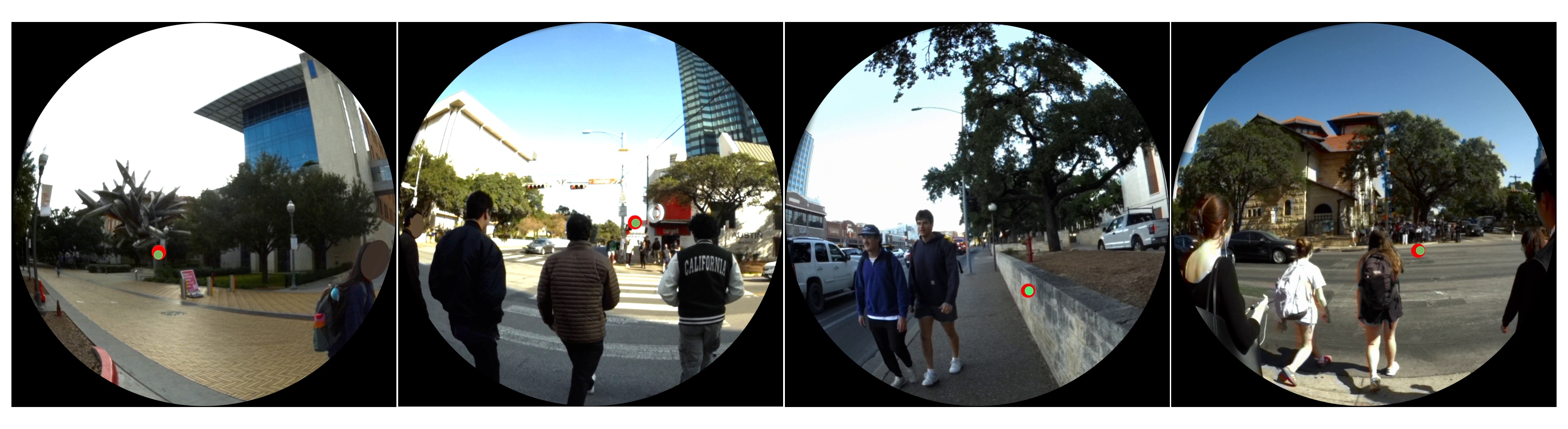}
    \caption{Gaze-to-pixel calibration examples. The green dot indicates the projected gaze fixation overlaid on egocentric video frames confirming accurate alignment between the 3D gaze stream and the egocentric video plane.}
  \label{fig:gaze-overlay}
\end{figure}

EgoTraj records per-frame 3D gaze origin and direction vector, capturing where the participant is looking relative to the walking scene.
Prior work in visual neuroscience has established that gaze anticipates locomotor actions by 1--2 seconds~\cite{land2006eye}.
For example, pedestrians often fixate on intended turning points, obstacles, or crossing opportunities well before executing the corresponding maneuver.
To associate 3D gaze geometry with the egocentric RGB stream, we fit a per-session quadratic calibration model that maps gaze yaw--pitch angles to pixel coordinates $(u, v)$ in the video frame.
This calibration enables predictive algorithms to directly project gaze fixation points into image space, as illustrated in \cref{fig:gaze-overlay}.
Combined with the scene annotations (\cref{sec:vlm}), the calibrated gaze stream supports joint reasoning about what pedestrians see and where they look, supporting intent-aware trajectory prediction for real world applications.

\begin{figure}[t]
  \centering
  \includegraphics[width=\linewidth]{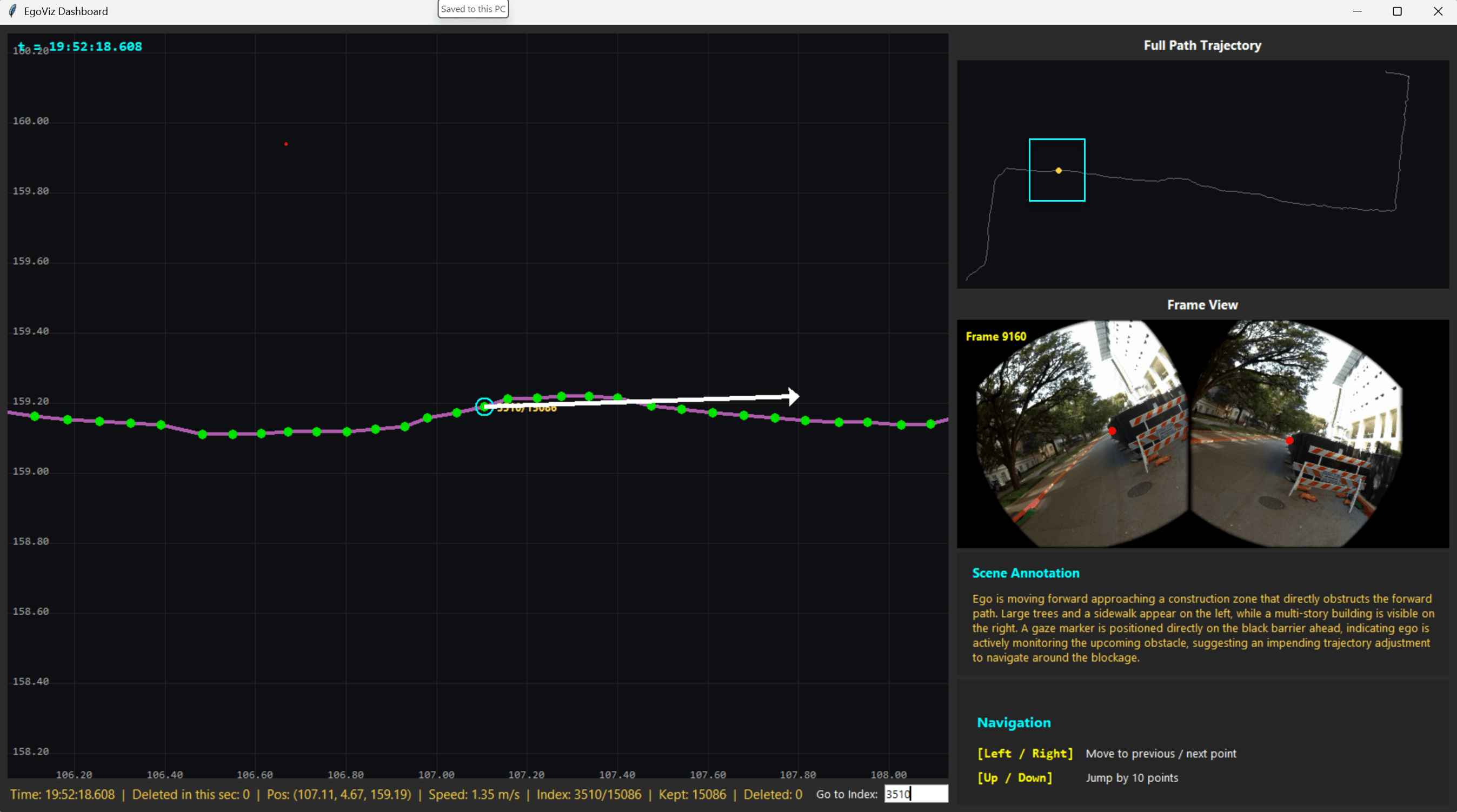}
    \caption{A snapshot of the EgoViz Dashboard showing synchronized trajectory, gaze, video, and annotation streams. \textbf{Left:} the 30\,Hz MQPro trajectory (green points) with the instantaneous gaze direction vector (white arrow) on a zoomable 2D map. \textbf{Top right:} a BEV of the full session path with the current position highlighted. \textbf{Middle right:} the corresponding egocentric RGB frame with projected gaze (red dot). \textbf{Bottom right:} VLM-generated scene description. The bottom status bar reports timestamp, 3D position, speed, and frame index for the selected sample.}
  \label{fig:egoviz}
\end{figure}

\subsection{Data Visualization and Quality Control}
To ensure synchronization consistency and annotation integrity across modalities, we developed \textit{EgoViz Dashboard}, an interactive inspection interface for the EgoTraj dataset (\cref{fig:egoviz}).
The dashboard synchronizes four complementary views:
(i) a zoomable 2D trajectory plot showing 30\,Hz pose samples with the corresponding gaze-direction vector,
(ii) a BEV of the full-session path,
(iii) the aligned egocentric RGB frame, and
(iv) the associated VLM-generated scene annotation.
Users can navigate recordings by timestamp or frame index, enabling frame-level verification of pose--gaze--video alignment.
The dashboard also displays session metadata including timestamp, 3D position, walking speed, and frame index to facilitate detailed inspection.
During data collection and preprocessing, the dashboard was used to detect synchronization drift, validate gaze reprojection accuracy, and identify sessions requiring reprocessing.
The \textit{EgoViz Dashboard} will be released alongside the dataset to support data exploration and reproducibility.

\section{Benchmarking}
\label{sec:benchmarking}

To validate EgoTraj as a benchmark for egocentric trajectory prediction, we evaluated several state-of-the-art baselines on trajectory forecasting, and performed ablation studies to quantify the predictive value of gaze and scene context.
All evaluations were conducted on a held-out test split with sessions unseen during training.
The results demonstrated that EgoTraj provides a challenging benchmark for egocentric navigation tasks and enabled systematic study of multimodal trajectory prediction.

\subsection{Quantitative Evaluation}
\label{sec:ego-traj-pred}

\paragraph{Metrics.}
For trajectory prediction, we reported Average Displacement Error (ADE) and Final Displacement Error (FDE).
ADE measures the mean Euclidean distance between predicted and ground-truth positions over all predicted timesteps, while FDE measures the Euclidean distance at the final predicted timestep.
For head orientation, we calculated the L1 rotation error~\cite{pan2025lookout}, computed as
\[
\mathcal{L}_\text{rot} =
\frac{1}{T_\text{pred}}
\sum_{i=1}^{T_\text{pred}}
\|\hat{R}_{t+i} R_{t+i}^{\top} - I\|_1,
\]
where $\hat{R}$ and $R$ denote the predicted and ground-truth rotation matrices, respectively.

\paragraph{Baselines.}
Egocentric trajectory prediction remains relatively underexplored, with few directly comparable baselines.
While single-agent trajectory models such as Social-LSTM~\cite{alahi2016social}, and TUTR~\cite{shi2023trajectory} exist, they are trained on BEV surveillance datasets and are not compatible with egocentric 6DoF pose sequences.
In addition, the closest related egocentric trajectory prediction works include EgoNav~\cite{wang2024egonav}, EgoCast~\cite{escobar2025egocast}, LookOut~\cite{pan2025lookout}, and EgoCogNav~\cite{qiu2025egocognav}.
However, EgoNav, LookOut, and EgoCogNav are currently available only as arXiv preprints without publicly released code.
Therefore, we adapted several representative methods that took multiple modalities simultaneously as their inputs:

\begin{itemize}
\renewcommand\labelitemi{$\bullet$}

\item \textbf{Constant Velocity} (\texttt{Const\_Vel})~\cite{pan2025lookout}, which extrapolates future body-frame translation and rotation using linear and angular velocities estimated from the last observed steps;

\item \textbf{Linear Extrapolation} (\texttt{Lin\_Ext})~\cite{pan2025lookout}, which fits a per-axis linear regression model to past translation and rotation observations and projects the motion sequence into the future;

\item \textbf{Multimodal Transformer} (\texttt{M\_Transformer})~\cite{pan2025lookout}, a standard Transformer baseline that takes the same inputs as the other learned baselines but performs early fusion by concatenating modality embeddings, followed by a temporal Transformer decoder and a linear head that predicts future trajectory;

\item \textbf{CXA-Transformer}~\cite{qiu2022egocentric}, a multi-stream cascaded cross-attention architecture with additional social and scene context streams. We adapt it by feeding the same ego-translation and rotation streams as \texttt{M\_Transformer}, but fusing them via cascaded cross-attention rather than simple concatenation, and

\item \textbf{EgoCast}~\cite{escobar2025egocast}, where we adapt its Transformer-based forecasting module, originally designed for full-body pose prediction, to take past ego-translation and head rotation as input and directly predict future trajectory, removing the pose-estimation stage.

\end{itemize}

All methods used an observation window of $T_\text{obs}=1.5$\,s to predict a future horizon of $T_\text{pred}=3.5$\,s, and were trained and evaluated using the same data split. All examined approaches were trained and evaluated using an 80/10/10 split of the EgoTraj sessions into training, validation, and test sets, respectively.

\begin{table}[t]
\centering
\caption{Baseline comparison on the EgoTraj test split. ADE and FDE are reported in meters; $L_1$\textsubscript{head} denotes the mean $L_1$ quaternion distance for head rotation prediction. Best results are in \textbf{bold}.}
\label{tab:main-results}
\setlength{\tabcolsep}{12pt}
\resizebox{0.6\linewidth}{!}{
\begin{tabular}{lccc}
\toprule
Model & ADE $\downarrow$ & FDE $\downarrow$ & $L_1$\textsubscript{head} $\downarrow$ \\
\midrule
Const\_Vel~\cite{pan2025lookout}        & 0.24 & 0.35 & 0.82 \\
Lin\_Ext~\cite{pan2025lookout}          & 0.26 & 0.39 & 1.39 \\
M\_Transformer~\cite{pan2025lookout}    & 0.20 & 0.32 & 0.74 \\
CXA-Transformer~\cite{qiu2022egocentric}   & 0.19 & 0.29 & \textbf{0.69} \\
EgoCast~\cite{escobar2025egocast}           & \textbf{0.16} & \textbf{0.28} & 0.78 \\
\bottomrule
\end{tabular}}
\end{table}

\paragraph{Results.}
Table~\ref{tab:main-results} reports quantitative results on ego-translation and rotation data modeling using the held-out EgoTraj test split.
Among all evaluated methods, the adapted EgoCast model achieves the best trajectory forecasting performance, obtaining the lowest ADE (0.16 m) and FDE (0.28 m).
In contrast, CXA-Transformer achieves the lowest head rotation error (0.69), indicating that its cascaded cross-attention mechanism better captures head orientation dynamics.
Classical kinematic baselines such as Constant Velocity and Linear Extrapolation perform reasonably for short-term translation but exhibit substantially higher rotation errors.
In particular, Linear Extrapolation yields the worst rotation accuracy (1.39 m), suggesting that simple per-axis regression fails to capture the nonlinear nature of head motion.
Neural sequence models consistently improve trajectory forecasting performance.
The vanilla Transformer baseline reduces ADE to 0.20 m and FDE to 0.32 m while improving rotation accuracy to 0.74, demonstrating the benefit of temporal modeling.
CXA-Transformer further improves both trajectory and rotation prediction compared to motion-based baselines, highlighting the advantage of cross-stream attention mechanisms in the egocentric setting.
Overall, these results indicate that learned temporal representations significantly outperform simple kinematic extrapolation for egocentric trajectory prediction, while multimodal attention mechanisms improve the modeling of head orientation.

\paragraph{Ablation Study.}
In addition to ego-translation and rotation, we performed an ablation study to analyze the contribution of different input modalities to egocentric trajectory prediction.
We adopted the CXA-Transformer~\cite{qiu2022egocentric} as the base architecture, since its cascaded cross-attention mechanism allowed progressive fusion of multiple modality streams.
We considered ego-motion $\mathcal{Y}$ (ego-translation and rotation), social context from nearby people represented as center points $\mathcal{C}$, bounding boxes $\mathcal{B}$, or body poses $\mathcal{P}$, scene understanding via semantic segmentation $\mathcal{S}$, relative depth $\mathcal{D}$, and gaze $\mathcal{G}$ represented as normalized image coordinates $(u,v)$.
Body pose $\mathcal{P}$, center points $\mathcal{C}$, and bounding boxes $\mathcal{B}$ were extracted using YOLOv8-Pose~\cite{yolov8ultralytics}, with center point $\mathcal{C}$ derived from the torso keypoint centroid.
Scene segmentation $\mathcal{S}$ was obtained from OneFormer~\cite{jain2023oneformer}.
Relative depth $\mathcal{D}$ was estimated using Depth Anything V2~\cite{yang2024depth}.
Gaze $\mathcal{G}$ was obtained directly from the calibrated gaze projection $(u,v)$.
Table~\ref{tab:ablation} summarizes the results.

\begin{table}[t]
\centering
\caption{Ablation study on input modalities using CXA-Transformer
on the EgoTraj test split.
$\mathcal{Y}$: ego-trajectory,
$\mathcal{C}$/$\mathcal{B}$/$\mathcal{P}$: nearby people
(center point / bounding box / pose),
$\mathcal{S}$: scene segmentation,
$\mathcal{D}$: relative depth,
$\mathcal{G}$: gaze $(u,v)$.
Best results per group in \textbf{bold}.}
\label{tab:ablation}
\setlength{\tabcolsep}{12pt}
{\small
\begin{tabular}{lccc}
\toprule
Modality & ADE $\downarrow$ & FDE $\downarrow$ & $L_1$\textsubscript{head} $\downarrow$ \\
\midrule
$\mathcal{Y}$~\cite{qiu2022egocentric}                                & 0.19 & 0.29 & 0.69 \\
\midrule
$\mathcal{Y} + \mathcal{C}$            & 0.18 & 0.29 & 0.79 \\
$\mathcal{Y} + \mathcal{B}$            & 0.18 & 0.30 & 0.81 \\
$\mathcal{Y} + \mathcal{P}$            & 0.17 & 0.27 & 0.77 \\
$\mathcal{Y} + \mathcal{S}$            & 0.16 & \textbf{0.26} & 0.74 \\
$\mathcal{Y} + \mathcal{D}$            & 0.18 & 0.29 & 0.78 \\
$\mathcal{Y} + \mathcal{G}$            & \textbf{0.15} & \textbf{0.26} & \textbf{0.69} \\
\midrule
$\mathcal{Y} + \mathcal{C} + \mathcal{G}$  & 0.14 & 0.25 & 0.67 \\
$\mathcal{Y} + \mathcal{B} + \mathcal{G}$  & 0.16 & 0.26 & 0.70 \\
$\mathcal{Y} + \mathcal{P} + \mathcal{G}$  & \textbf{0.12} & \textbf{0.24} & \textbf{0.63} \\
$\mathcal{Y} + \mathcal{S} + \mathcal{G}$  & \textbf{0.12} & 0.25 & 0.65 \\
$\mathcal{Y} + \mathcal{D} + \mathcal{G}$  & 0.15 & 0.27 & 0.71 \\
\midrule
$\mathcal{Y} + \mathcal{P} + \mathcal{S} + \mathcal{G}$
                                            & \textbf{0.12} & \textbf{0.23} & \textbf{0.58} \\
\bottomrule
\end{tabular}
}
\end{table}

Starting from ego-motion only ($\mathcal{Y}$), both scene segmentation ($\mathcal{S}$) and gaze ($\mathcal{G}$) provide the largest individual improvements. Adding scene context reduces ADE from 0.19 to 0.16, while incorporating gaze further reduces it to 0.15. This suggests that semantic layout and visual attention both contribute anticipatory information beyond kinematic motion cues.
Among social representations, pose ($\mathcal{P}$) consistently outperforms bounding boxes ($\mathcal{B}$) and center points ($\mathcal{C}$), achieving the lowest trajectory error among the three. This finding suggests that fine-grained human motion cues are more informative than coarse spatial localization for modeling social interactions.
In addition, adding gaze to social and scene features consistently improves performance across configurations.
The full multimodal combination ($\mathcal{Y}+\mathcal{P}+\mathcal{S}+\mathcal{G}$) achieves the best overall performance, reducing ADE to 0.12, FDE to 0.23, and head rotation error to 0.58.
These results highlight that visual attention provides additional predictive signal even in the presence of rich scene and social context.

\paragraph{Generalization across splits.}
To examine how well the multimodal models generalize beyond the random-participant split, we evaluated CXA-Transformer on two additional, stricter splits.
The \textit{waypoint-pair held-out split} reserves 3 of the 21 origin--destination pairs entirely for testing ($n{=}10$ sessions), and the \textit{unfamiliar split} reserves 8 of the 31 participants who reported being unfamiliar with the recording environment ($n{=}8$ sessions).
By construction, each session is contributed by a unique participant (Sec.~\ref{sec:dataset}), so all three splits are subject-disjoint.
Table~\ref{tab:splits} reports ADE/FDE on each split with 95\% bootstrap confidence intervals computed from 1000 resamples.
The full multimodal configuration ($\mathcal{Y}{+}\mathcal{P}{+}\mathcal{S}{+}\mathcal{G}$) remains the strongest across all three splits, with a modest generalization gap (random-participant ADE $0.12 \to$ waypoint-held-out ADE $0.14 \to$ unfamiliar ADE $0.14$).
This indicates that the multimodal cues transfer to held-out landmark pairs and to participants unfamiliar with the area, rather than overfitting to specific route templates.

\begin{table}[t]
\centering
\caption{Generalization across three test splits using CXA-Transformer. Values are ADE/FDE in meters with 95\% bootstrap confidence intervals from 1000 resamples. Best per split in \textbf{bold}.}
\label{tab:splits}
\setlength{\tabcolsep}{4pt}
\renewcommand{\arraystretch}{1.1}
\newcommand{\ci}[1]{{\scriptsize$\,\pm$#1}}
{\small%
\begin{tabular}{lcccccc}
\toprule
 & \multicolumn{2}{c}{\textbf{Random Participant} ($n{=}8$)} & \multicolumn{2}{c}{\textbf{Waypoint Held-Out} ($n{=}10$)} & \multicolumn{2}{c}{\textbf{Unfamiliar} ($n{=}8$)} \\
\cmidrule(lr){2-3}\cmidrule(lr){4-5}\cmidrule(lr){6-7}
Modality & ADE $\downarrow$ & FDE $\downarrow$ & ADE $\downarrow$ & FDE $\downarrow$ & ADE $\downarrow$ & FDE $\downarrow$ \\
\midrule
$\mathcal{Y}$                                & 0.19\ci{.014} & 0.29\ci{.021} & 0.21\ci{.018} & 0.32\ci{.024} & 0.23\ci{.019} & 0.34\ci{.027} \\
$\mathcal{Y}+\mathcal{P}$                    & 0.17\ci{.011} & 0.27\ci{.019} & 0.19\ci{.015} & 0.29\ci{.022} & 0.20\ci{.013} & 0.31\ci{.020} \\
$\mathcal{Y}+\mathcal{S}$                    & 0.16\ci{.013} & 0.25\ci{.014} & 0.18\ci{.012} & 0.28\ci{.018} & 0.18\ci{.016} & 0.29\ci{.023} \\
$\mathcal{Y}+\mathcal{G}$                    & 0.15\ci{.009} & 0.26\ci{.017} & 0.16\ci{.014} & 0.26\ci{.013} & 0.16\ci{.010} & 0.29\ci{.018} \\
$\mathcal{Y}+\mathcal{P}+\mathcal{S}+\mathcal{G}$ & \textbf{0.12}\ci{.008} & \textbf{0.23}\ci{.012} & \textbf{0.14}\ci{.010} & \textbf{0.25}\ci{.011} & \textbf{0.14}\ci{.012} & \textbf{0.26}\ci{.014} \\
\bottomrule
\end{tabular}
}
\end{table}

\subsection{Qualitative Evaluation}
\label{sec:qualitative}

\begin{figure}[t]
\centering
\includegraphics[width=\textwidth]{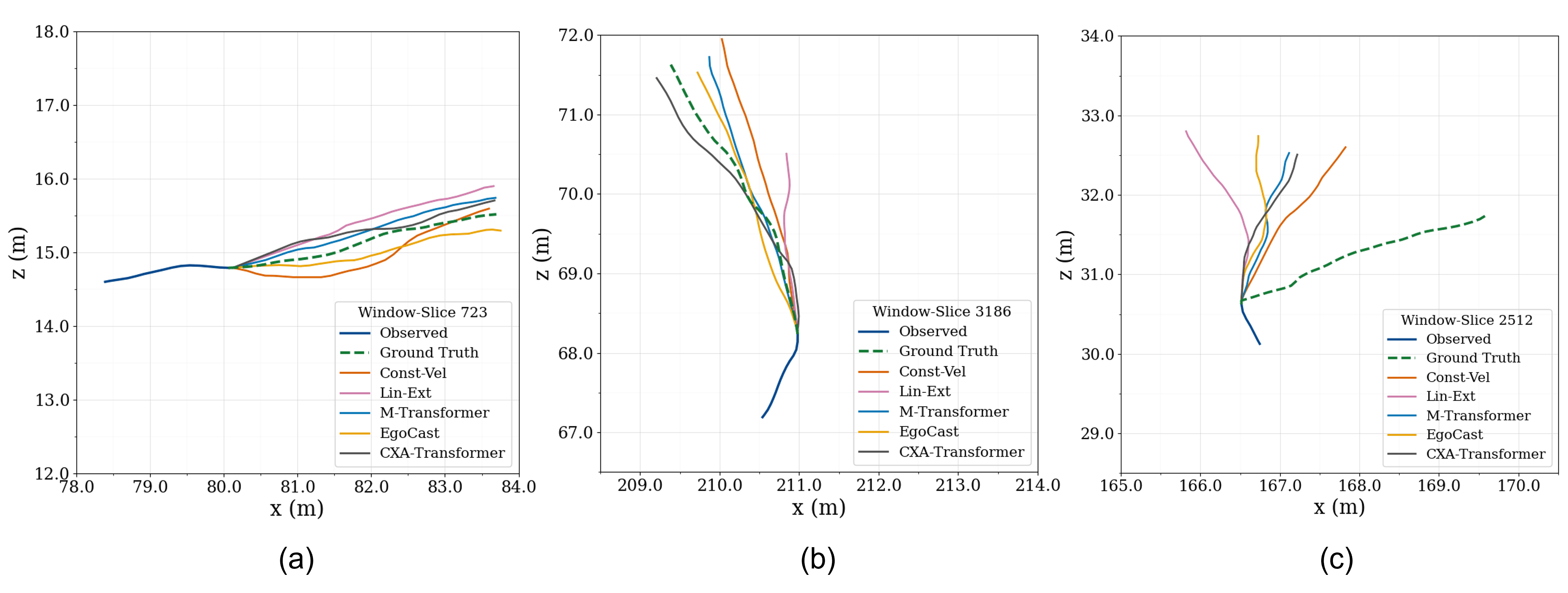}
\caption{\textbf{Qualitative trajectory forecasting.}
Egocentric trajectory predictions from multiple baselines using motion data (ego-translation and rotation) on three scenarios from the EgoTraj test split.
Dark blue: observed path; green dashed: ground truth; colors: predictions.
\textit{Left:} gentle segment.
\textit{Center:} moderate turn where attention-based models better follow the trajectory.
\textit{Right:} sharp $\sim 90^{\circ}$ intersection turn where all baselines underestimate the turning magnitude.}
\label{fig:egotraj_qual}
\end{figure}

Figure~\ref{fig:egotraj_qual} presents representative trajectory forecasting examples across diverse navigation scenarios.
Compared to motion-based baselines, multimodal models better capture turning behavior and obstacle avoidance.
In particular, CXA-Transformer and EgoCast models produce smoother and more socially compliant trajectories that align closely with ground truth, while kinematic baselines often overshoot or fail to anticipate directional changes.

\begin{figure}[H]
\centering
\includegraphics[width=\textwidth]{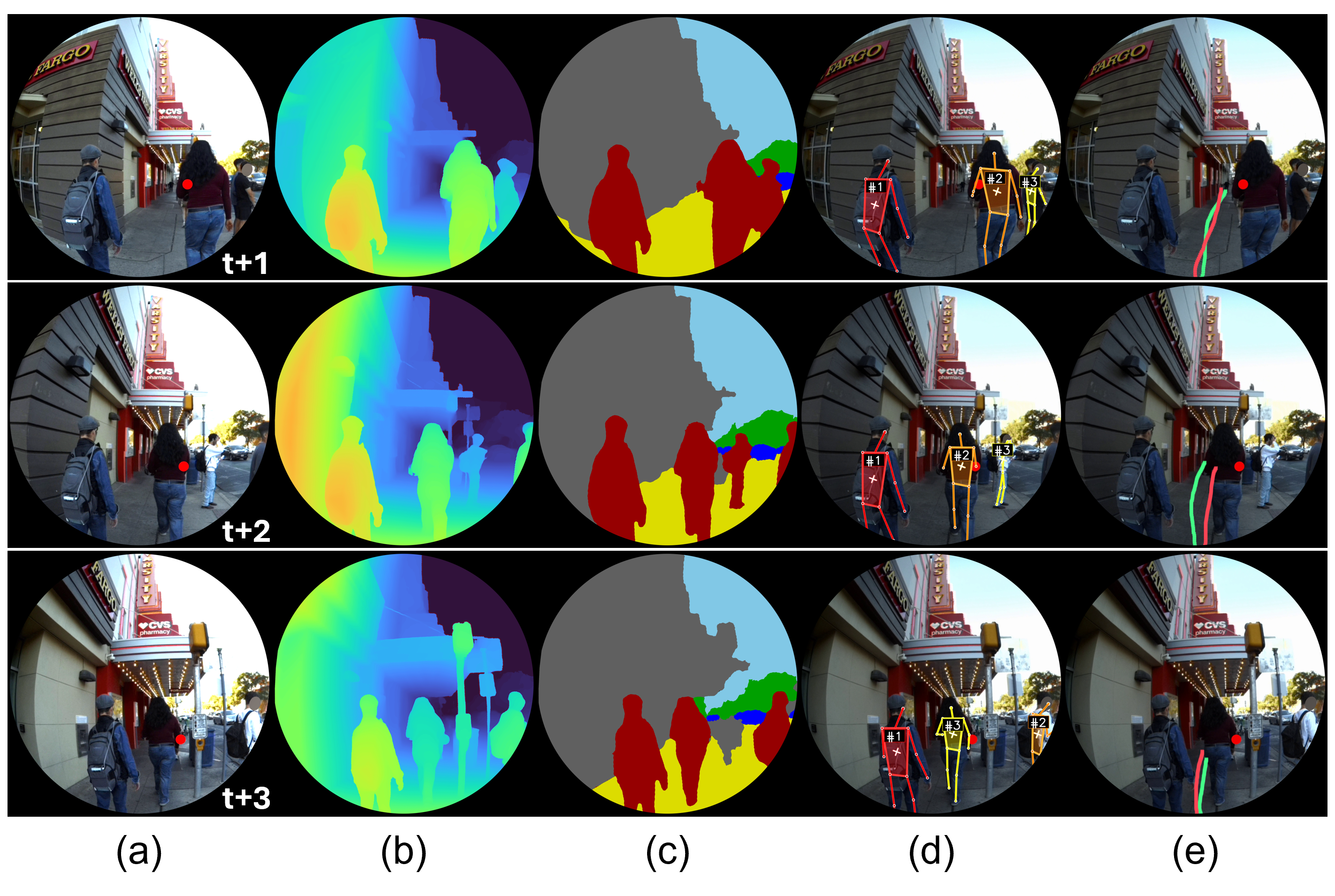}
\caption{\textbf{Multimodal observations across three consecutive timesteps.}
Each row corresponds to a frame at time $t{+}1$, $t{+}2$, and $t{+}3$
during a sidewalk navigation sequence.
From left to right: egocentric RGB frame with gaze fixation (red dot),
relative depth estimated by Depth Anything V2,
semantic segmentation predicted by OneFormer,
nearby pedestrians detected by YOLOv8-Pose ranked by depth proximity, and ground-truth (green) versus predicted (red) future trajectory
projected onto the egocentric frame.}
\label{fig:egotraj-multimodal}
\end{figure}

Figure~\ref{fig:egotraj-multimodal} visualizes the corresponding multimodal cues for CXA-Transformer at consecutive timesteps ($t{+}1$ to $t{+}3$), including relative depth, semantic segmentation, detected human pose, and projected gaze.
Relative depth encodes the proximity of surrounding objects to the wearer, while scene segmentation provides structural priors such as walkable regions and obstacles.
Human pose captures nearby pedestrian motion, and gaze indicates the wearer's visual attention toward potential navigation targets.
The predicted trajectories of full multimodal combination ($\mathcal{Y}+\mathcal{P}+\mathcal{S}+\mathcal{G}$) closely follow these attended regions, illustrating the complementary role of visual attention in guiding motion forecasting.
Overall, these qualitative results demonstrate that jointly modeling ego-motion, scene structure, social context, and gaze leads to more realistic and anticipatory trajectory
predictions in complex urban environments.

\begin{figure}[H]
\centering
\setlength{\abovecaptionskip}{2pt}
\includegraphics[width=0.37\linewidth]{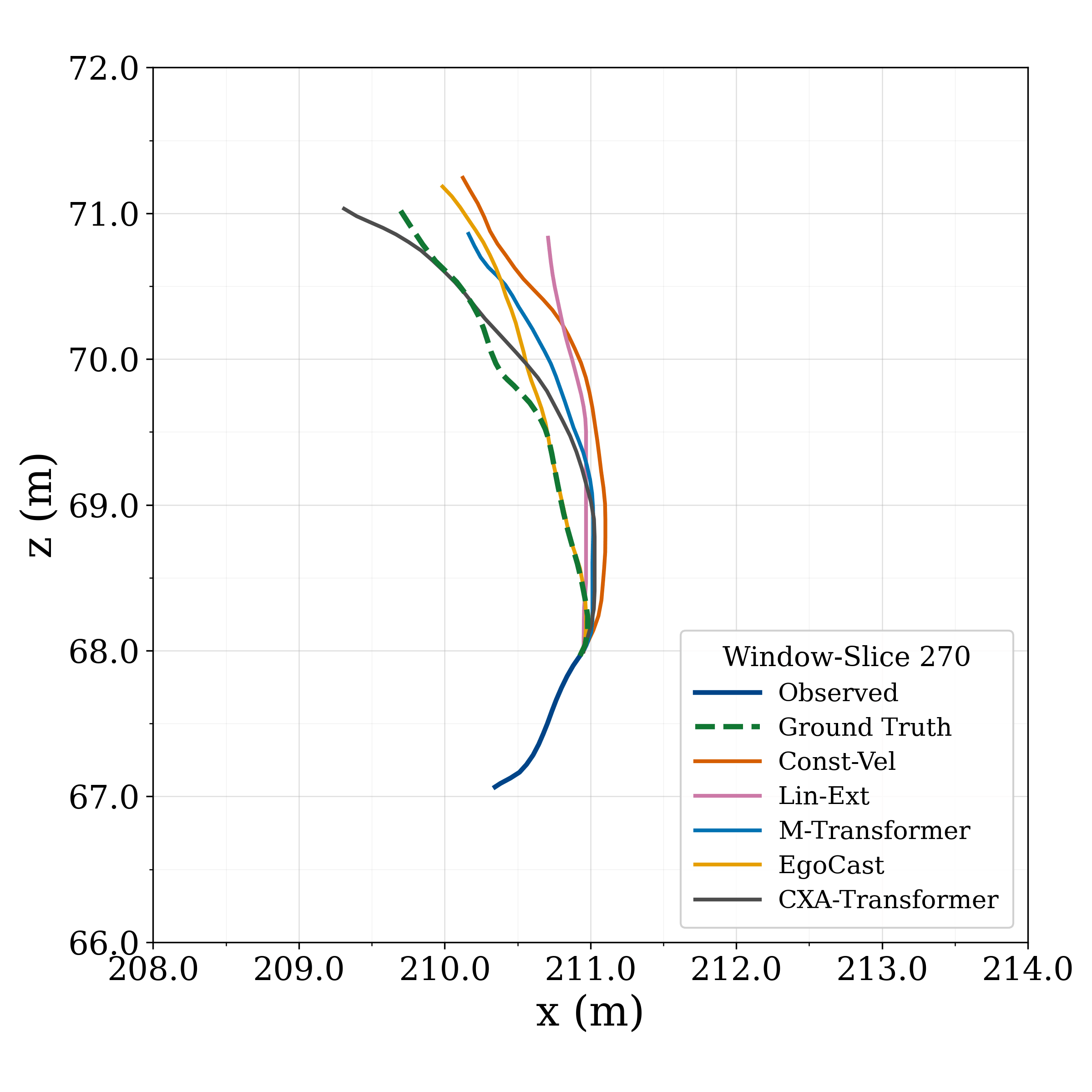}\hspace{0.02\linewidth}\includegraphics[width=0.37\linewidth]{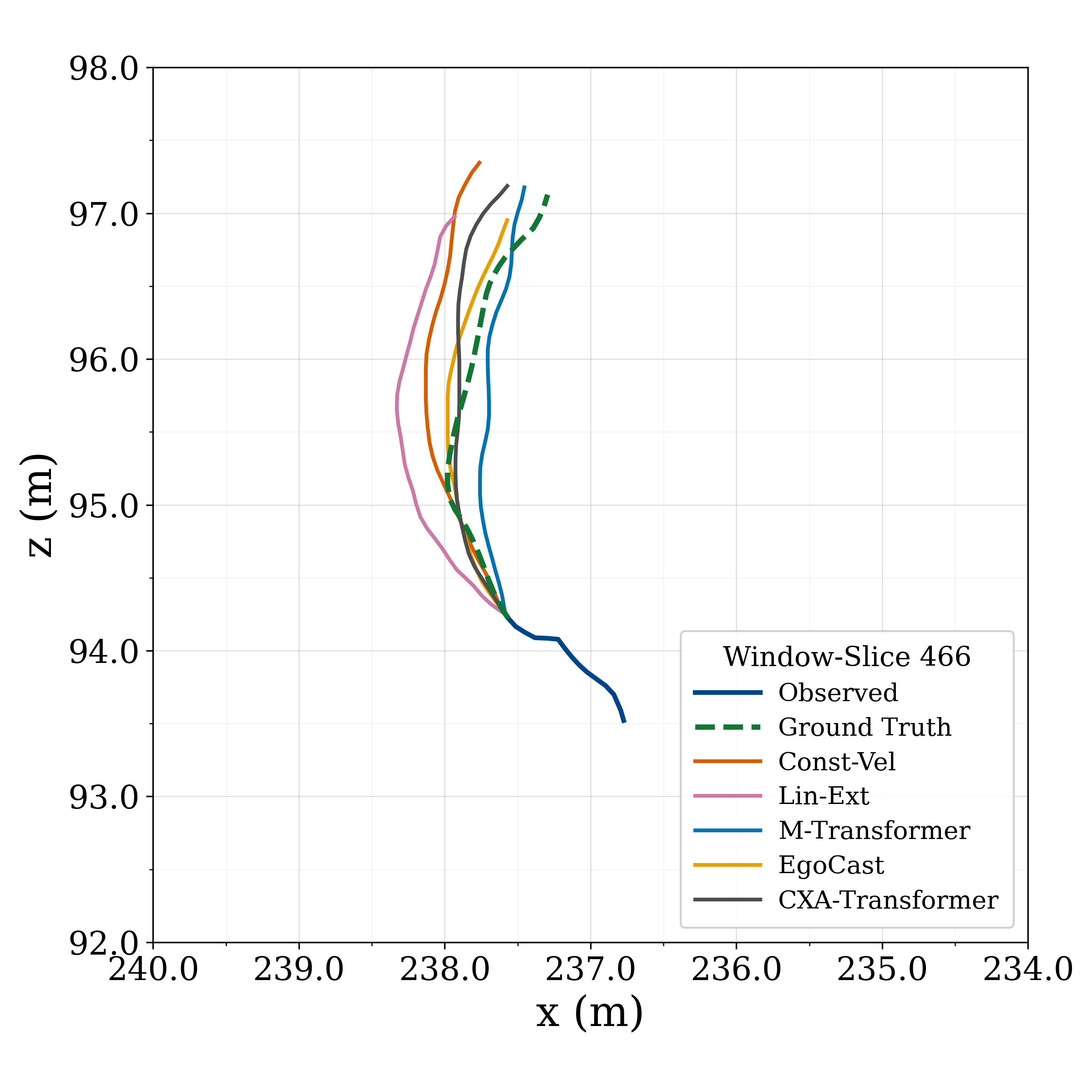}
\caption{\textbf{Active-transition windows} where turning begins within the final $0.5$\,s of $T_\text{obs}$. Multimodal transformer-based predictors (CXA-Transformer, EgoCast) track the ground-truth trajectory through the turn, while motion-only baselines (Const-Vel, Lin-Ext, position-only) drift outward along the pre-turn heading, consistent with gaze leading motion by 1--2\,s before turns.}
\label{fig:active-transition}
\end{figure}

\paragraph{Failure Cases.}
Figure~\ref{fig:egotraj_qual} (Window-Slice 2512) illustrates a representative failure case involving a sharp $\sim 90^{\circ}$ turn at an intersection.
After noticing the traffic light turn green, the participant abruptly changes heading within a short temporal window, causing all baselines to deviate from the ground-truth trajectory.
Classical kinematic models continue extrapolating along the pre-turn heading, while learned models partially anticipate the direction change but still underestimate the
true curvature of the maneuver.
This failure arises from the limited predictive signal available in short-horizon ego-motion history. Prior to the turn, both linear and angular velocities remain within normal walking ranges, providing little indication that a sharp maneuver is imminent.
In addition, the dataset did not include enough data representing such abrupt transitions.
More broadly, this behavior reflects an inherent challenge of deterministic trajectory forecasting under multimodal intent.
Abrupt turns at intersections are difficult to predict without strong anticipatory cues or richer environmental context, motivating future research on intent-aware and uncertainty-aware egocentric prediction.

\paragraph{Anticipation under Active Transitions.}
A central motivation for incorporating gaze is that visual attention precedes locomotor action by approximately 1--2\,s~\cite{land2006eye}, especially around turns.
To isolate this regime, we identified ``active transition'' windows in which turning behavior begins within the final $0.5$\,s of the observation window $T_\text{obs}$, so most of the directional change occurs in the prediction horizon rather than the observed past.
On this subset, the full multimodal CXA-Transformer achieves ADE $0.23 \pm 0.022$ and FDE $0.38 \pm 0.028$.
Figure~\ref{fig:active-transition} shows two representative active-transition windows: multimodal transformer-based models track the ground-truth trajectory through the turn, while motion-only baselines (Const-Vel, Lin-Ext, position-only) drift outward along the pre-turn heading.
This pattern is consistent with the hypothesis that gaze provides anticipatory cues before motion-based features express the upcoming maneuver.
Abrupt, near-instantaneous sharp turns remain challenging due to the limited number of such transitions in the dataset; we leave this regime to future work.

\section{Conclusion}
\label{sec:conclusion}

In this paper, we introduce EgoTraj, a large-scale multimodal egocentric trajectory dataset collected from 75 sessions using Meta Quest Pro headsets in real-world urban environments. To the best of our knowledge, EgoTraj is the first large-scale egocentric trajectory dataset that jointly provides synchronized 6DoF head pose, per-frame 3D gaze vectors, egocentric RGB video, and scene annotations recorded in real traffic scenarios.
Through benchmark experiments on several state-of-the-art trajectory forecasting models, we showed that EgoTraj presents a challenging benchmark for egocentric prediction, reflecting distinctive motion patterns such as frequent turns and abrupt directional changes. In addition, ablation studies demonstrated that gaze and scene cues provide complementary predictive signals beyond ego-motion alone, highlighting the importance of visual attention in anticipating human navigation intent. Finally, we release EgoTraj together with scene annotations and the EgoVis dashboard to support future research on egocentric trajectory prediction, and vision--language understanding of navigation intent.

\section*{Acknowledgements}

This project was carried out as a collaborative research effort between Honda Development \& Manufacturing of America, LLC (HDMA) and MASSlab at The University of Texas at Austin, focused on multimodal egocentric perception and trajectory prediction. The project aims to better understand human navigation behavior to improve human--vehicle interactions in traffic environments and enhance the safety of vulnerable road users.

We thank Prof.~Kristen Grauman (The University of Texas at Austin) for insightful discussions and constructive feedback throughout this work.

\bibliographystyle{splncs04}
\bibliography{main_reference}

\clearpage
\appendix
\titleformat{\section}{\LARGE\bfseries}{\thesection}{1em}{}
\renewcommand{\thesection}{Appendix~\Alph{section}}
\renewcommand{\thesubsection}{\Alph{section}.\arabic{subsection}}

\section{Capture Setup and Recording Details}
\label{sec:appendix-capture}

All data was collected using the Meta Quest Pro headset (MQPro), a mixed-reality device equipped with integrated eye-tracking cameras, inside-out positional tracking, and a front-facing RGB passthrough camera. The device performs head tracking using a combination of inertial measurement units (IMU) and four wide-angle monochrome cameras that enable real-time visual--inertial odometry.

\subsection{Recording Profile}
During data collection, the headset records egocentric RGB video at 30\,fps with a resolution of $1024 \times 1024$ pixels.
The system simultaneously logs the wearer's 6DoF head pose, including 3D position and orientation represented as quaternions, and binocular gaze direction vectors at 50\,Hz.
All sensor streams share timestamps generated by the headset's internal clock and are exported with nanosecond precision.
To obtain a unified multimodal representation, pose and gaze measurements were temporally aligned with the video frame rate during preprocessing pipeline.
The final per-frame dataset representation includes the RGB frame, head position $(x,y,z)$, head orientation $(q_w,q_x,q_y,q_z)$, and left/right 3D gaze direction vectors.

\subsection{Recording Pipeline}
\label{sec:appendix-pipeline}

\begin{figure}[t]
\centering
\includegraphics[width=0.45\textwidth]{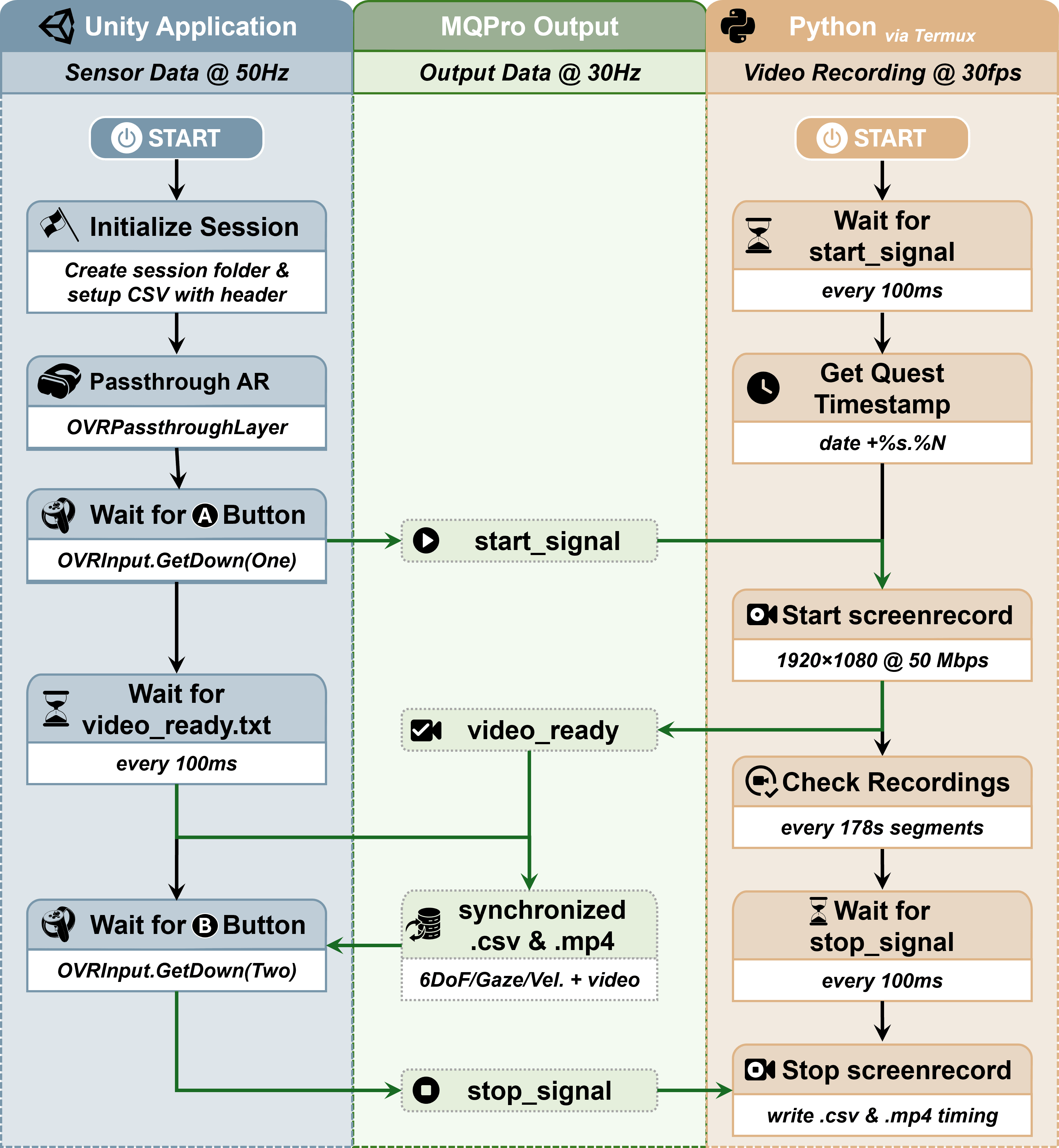}
\caption{
Overview of the multimodal recording pipeline used to collect the EgoTraj dataset.
A custom Unity application running on the MQPro logs sensor data at 50\,Hz,
including head pose and gaze measurements. A Python-based recording script captures
egocentric RGB video at 30\,fps. Synchronization between the two processes is achieved
through file-based signals (\texttt{start\_signal}, \texttt{video\_ready}, and
\texttt{stop\_signal}), ensuring temporally aligned sensor streams.
}
\label{fig:recording-pipeline}
\end{figure}

To capture synchronized multimodal data streams, we developed a custom recording pipeline combining a Unity application with a Python-based recording script running on the headset.
Figure~\ref{fig:recording-pipeline} illustrates the full data capture workflow.
The Unity application logs sensor measurements at 50\,Hz, including head pose, gaze vectors, and motion signals such as linear and angular velocity.
At the beginning of each session, the application initializes the recording directory, creates the sensor logging files, and waits for user input to start the recording.
When recording begins, the Unity application emits a \texttt{start\_signal} trigger that activates a Python-based screen recording process running in the headset environment.
The Python script records egocentric RGB video at 30\,fps and retrieves device timestamps to ensure temporal consistency with the sensor streams.
A synchronization handshake between the two processes is implemented using file-based signals (\texttt{start\_signal}, \texttt{video\_ready}, and \texttt{stop\_signal}).
Once the video recording process confirms readiness, the Unity application continues logging sensor data while the video stream is captured.
The temporal alignment between modalities is bounded by the frame interval (approximately $5.893$\,ms).
At the end of the session, both processes terminate simultaneously and export temporally aligned sensor logs and video recordings.

\subsection{Recording Procedure}
\label{sec:appendix-procedure}

Each recording session follows a standardized collection protocol:

\begin{enumerate}
\item Initialize the Unity recording application on the Meta Quest Pro headset.
\item The application creates a new session directory and prepares the sensor logging files.
\item The participant wears the headset and selects the two waypoints in the recording environment.
\item The participant triggers the start of the session using the controller A button, which activates the \texttt{start\_signal} and begins both sensor logging and video recording.
\item During the session, the headset's wearer navigates naturally through the environment while the system records synchronized RGB video, head pose, and gaze measurements.
\item When the recording is complete, the participant triggers the controller B button, which sends the \texttt{stop\_signal} and terminates both processes.
\item The recorded data are exported as synchronized sensor logs and video files for subsequent preprocessing.
\end{enumerate}

\subsection{Data Processing}
\label{sec:appendix-postprocess}

\begin{figure}[!t]
\centering
\includegraphics[width=0.4\textwidth]{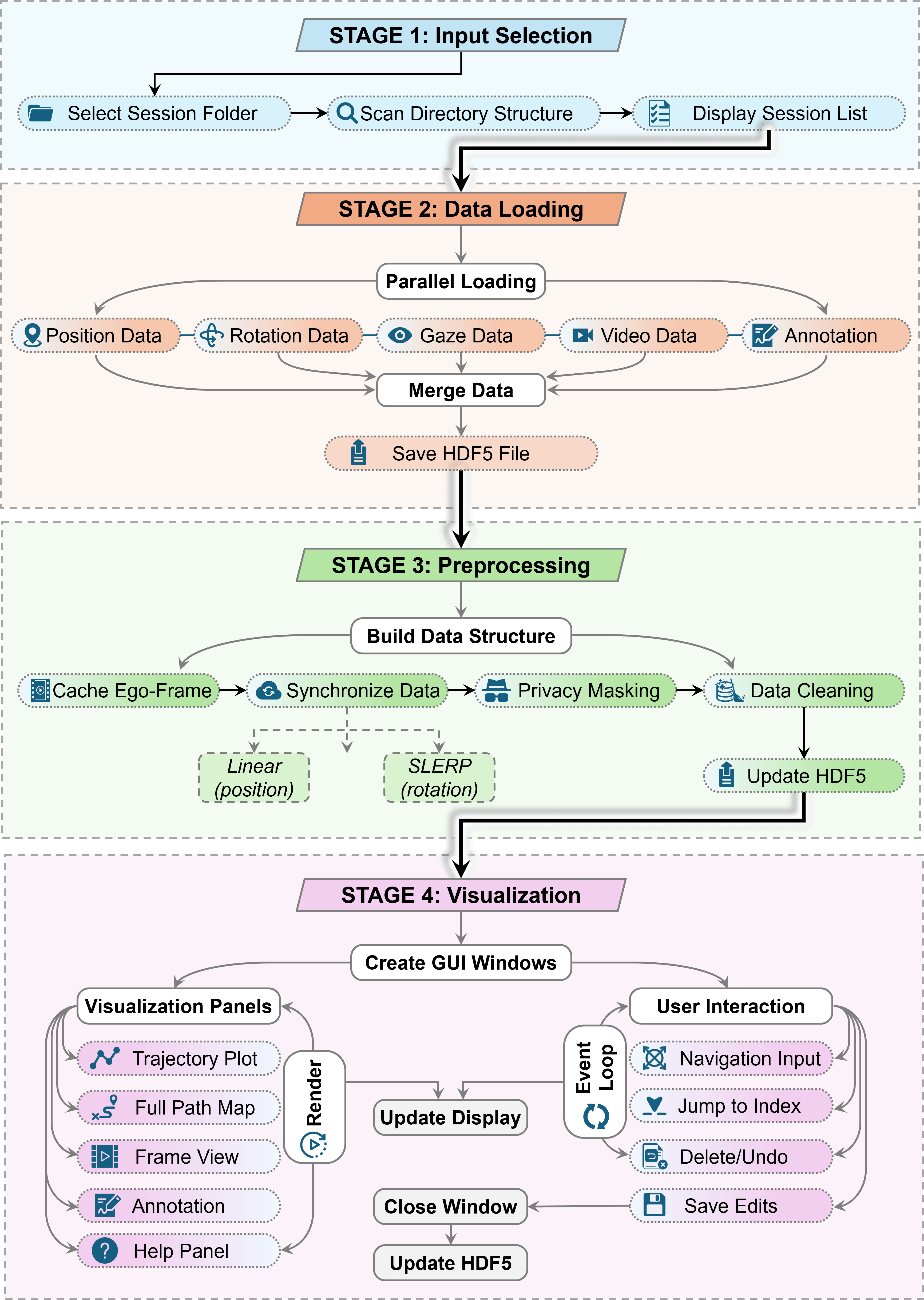}
\caption{
Overview of the EgoTraj preprocessing pipeline. Stage 1 selects sessions and scans the dataset structure. Stage 2 loads multimodal streams (position, rotation, gaze, RGB video, and annotations), merges them, and creates the initial HDF5 dataset. Stage 3 synchronizes and preprocesses the data using interpolation (linear for position and SLERP for rotations), privacy masking, and data cleaning to produce processed HDF5. Stage 4 provides interactive visualization tools for inspecting trajectories, browsing frames with annotation, and validating multimodal recordings.
}
\label{fig:data-preprocessing}
\end{figure}

After data collection, the recorded multimodal streams were processed using a custom preprocessing pipeline designed to synchronize and organize the data into a unified dataset format. The preprocessing workflow is illustrated in Figure~\ref{fig:data-preprocessing}.

\noindent\textbf{Input selection.}
The preprocessing pipeline began by selecting a recording session directory containing the raw sensor logs and video recordings. The directory structure was automatically scanned to identify available session files and metadata. Each session contained RGB video streams, timestamp logs, head pose measurements, gaze vectors, and motion signals.

\noindent\textbf{Data loading.}
All modalities were loaded in parallel to improve processing efficiency. The pipeline read the raw sensor streams and merges them into a unified dataset structure before storing them in an HDF5 container. These streams were merged using their shared timestamps and stored as the initial HDF5 dataset.

\noindent\textbf{Preprocessing and synchronization.}
Because sensors operated at different sampling rates (\eg, gaze at 50\,Hz and RGB video at 30\,fps), all modalities were aligned to a common timeline corresponding to the video frame rate. Synchronization was performed using the device timestamps generated by the MQPro internal clock. Missing samples were resolved through modality-specific interpolation. Linear interpolation was applied to position signals, spherical linear interpolation (SLERP) was used for quaternion-based head rotations, and normalized linear interpolation was applied to gaze vectors. These steps ensured smooth and temporally consistent trajectories across modalities.

\noindent\textbf{Data cleaning and privacy filtering.}
After synchronization, the dataset underwent automated validation and filtering. Sessions with incomplete recordings or inconsistent sensor logs were removed to ensure temporally coherent multimodal sequences. Identifiable visual elements were masked when necessary to protect participant privacy. Selected sequences checks were manually inspected to confirm correct temporal alignment between all recorded multimodal dataset.

\noindent\textbf{Visualization and inspection tools.}
To facilitate dataset inspection and debugging, we developed \textit{EgoVis}, an interactive visualization tool that loads the processed dataset and displays synchronized multimodal streams. The interface provided trajectory plots, full path maps, frame-level RGB inspection, and timeline navigation. Users could browse sequences, validate synchronization, and export corrected data back to the HDF5 dataset. This pipeline ensures high-quality, temporally aligned observations for trajectory prediction and multimodal perception research.

\section{Participants}
\label{sec:appendix-participants}

Participants volunteered to take part in the data collection sessions.
All participants provided informed consent prior to recording and were informed about the purpose of the dataset collection.
Participant surveys were separated into two questionnaires: a pre-session questionnaire and a post-session questionnaire, as shown in Table~\ref{tab:participant-survey}.
The pre-session questionnaire captures background information about the participant and their familiarity with the AR headsets and recording environment.
The post-session questionnaire collects feedback regarding the recording experience, including comfort while wearing the headset and perceived difficulty of the navigation task.
Both questionnaires were designed to be lightweight and quick to complete, as participants typically filled them out immediately before and after the recording session.
The survey consisted primarily of multiple-choice and Yes/No questions, with optional feedback.

\begin{table}[t]
\centering
\small
\caption{Pre-session and post-session questionnaire items.}
\label{tab:participant-survey}
\begin{tabular}{p{0.75\textwidth}p{0.18\textwidth}}
\toprule
\textbf{Question} & \textbf{Type} \\
\midrule
\multicolumn{2}{l}{\textit{Pre-Session}} \\
How familiar are you with the recording environment? & Multiple choice \\
How comfortable do you feel walking with an AR headset? & Multiple choice \\
Do you have any issue that prevent you from using AR headsets? & Yes / No \\
Have you previously used AR/VR head-mounted devices? & Yes / No \\
Have you participated in similar navigation experiments before? & Yes / No \\
How long do you expect the navigation session to take? & Free text \\

\midrule
\multicolumn{2}{l}{\textit{Post-Session}} \\
Self-reported navigation difficulty & Multiple choice \\
Did you complete the intended route? & Yes / No \\
Did the headset affect your natural walking behavior? & Yes / No \\
How comfortable was the headset during the session? & Multiple choice \\
How easy was the recording start and stop process? & Multiple choice \\
Did you experience any issues while navigating? Any additional comments?  & Free text \\
\bottomrule
\end{tabular}
\end{table}

\section{Area of Interest}
\label{sec:appendix-area}

\begin{figure}[t]
\centering
\includegraphics[width=0.5\textwidth]{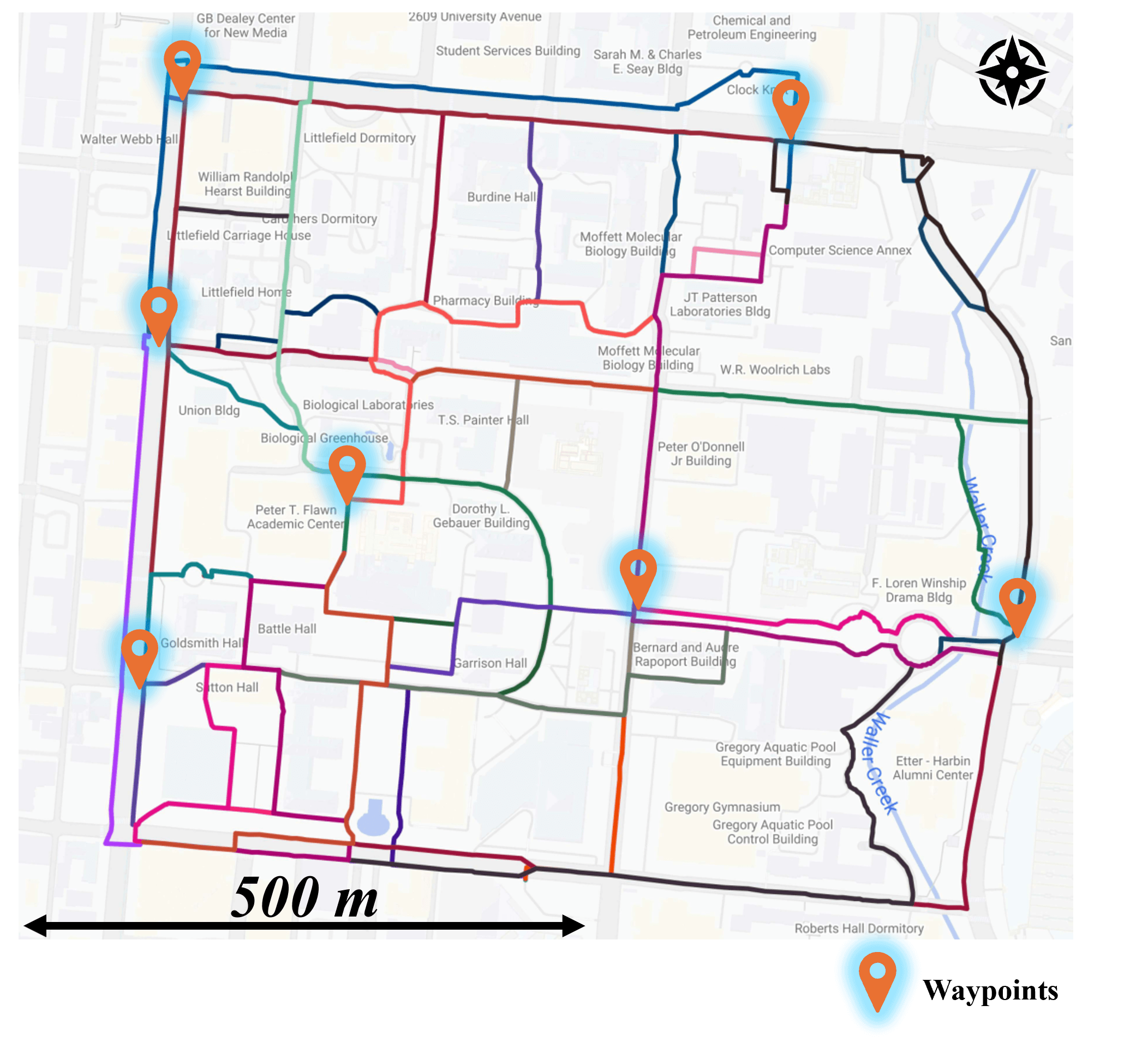}
\caption{Area of interest for EgoTraj data collection. Colored lines denote recorded walking routes, and markers indicate the seven entry/exit waypoints.}
\label{fig:area-map}
\end{figure}

Data collection was conducted in-the-wild traffic environment.
The selected region covered a dense network of pedestrian walkways connecting commercial buildings, university campus, food courts, and public services.
In particular, the region included multiple signalized and stop intersections, high-density pedestrian zones, curved paths, and narrow corridors between buildings, which naturally induce directional changes and trajectory variations.
Public transportation infrastructure was also integrated into the region with total 34 bus stops. These structural characteristics made the environment well suited to capture realistic human navigation behavior under a variety of spatial and social conditions. Figure~\ref{fig:area-map} illustrates the layout of the recording area and representative navigation routes.

\section{Scene Annotation Generation}
\label{sec:appendix_scene_annotation}

To augment EgoTraj with high-level contextual information, we generated structured scene annotations using Qwen2.5-VL-7B. The annotations describe navigation-relevant elements observed in egocentric frames, including surrounding pedestrians, traffic activity, environmental layout, and gaze fixation targets. Unlike generic image captions, the generated descriptions emphasize cues relevant to human navigation behavior.

\subsection{Chain-of-Thought Annotation Strategy}

To improve consistency and reasoning quality, we used a chain-of-thought (CoT) prompting strategy when generating annotations. Instead of producing a single caption directly, the model was guided to analyze the scene through a sequence of intermediate reasoning steps. This structured prompting encouraged the model to explicitly consider visual context, dynamic agents, and user attention before producing the final description. Specifically, the prompt instructed the model to reason about the scene in the following stages:

\begin{figure}[t]
\centering
\includegraphics[width=0.9\textwidth]{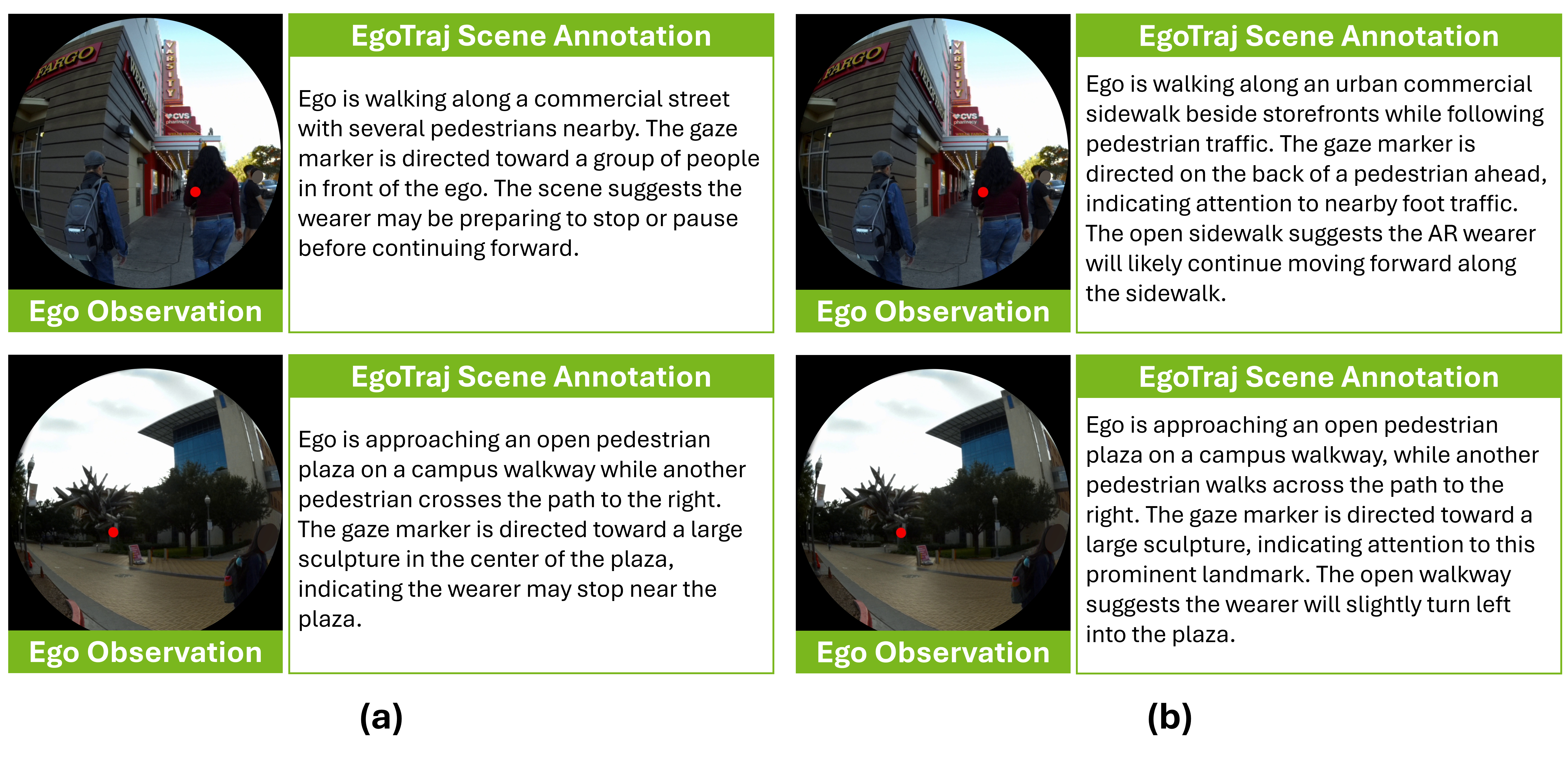}
\caption{Examples of EgoTraj scene annotations. Each observation frame shows the gaze marker (red dot) and the corresponding VLM-generated description. (a) initial annotations, and (b) refined annotations generated using the chain-of-thought prompting strategy.}
\label{fig:append_scene_annotations}
\end{figure}

\begin{enumerate}
\item Identify the environmental context (\eg, crosswalk, sidewalk, intersection).
\item Detect nearby dynamic agents such as pedestrians, vehicles, or cyclists.
\item Analyze traffic signals, obstacles, or navigation constraints.
\item Incorporate the projected gaze as an indicator of the user's attention.
\item Infer the likely short-term motion or navigation intent of the camera wearer.
\end{enumerate}

\subsection{Annotation Quality Evaluation}

To verify the reliability of the generated annotations (Figure~\ref{fig:append_scene_annotations}), we evaluated the pipeline using several complementary metrics on the same 100-frame stratified sample.
Structural compliance measured whether each annotation follows the predefined schema and reasoning structure.
For semantic quality, we report per-field Cohen's $\kappa$ between two human annotators and the VLM accuracy against the human reference for each of the five structured fields: environmental context, dynamic agents, traffic signals, gaze target, and short-term intent.
Table~\ref{tab:annotation_quality} summarizes the results.
Inter-annotator agreement ranges from $\kappa=0.83$ on the inferential \emph{short-term intent} field to $\kappa=0.96$ on observable \emph{traffic signals}, with VLM accuracy tracking annotator agreement closely.
Structural compliance over the same sample reached $96\%$ after up to two prompt retries.

\begin{table}[t]
\centering
\small
\caption{Per-field annotation quality on the 100-frame stratified sample. Cohen's $\kappa$ is computed between two human annotators, and VLM accuracy is against the human reference.}
\label{tab:annotation_quality}
\setlength{\tabcolsep}{8pt}
\begin{tabular}{lcc}
\toprule
\textbf{Annotation field} & \textbf{Cohen's $\kappa$} & \textbf{VLM accuracy} \\
\midrule
Environmental context & 0.89 & 0.91 \\
Dynamic agents        & 0.87 & 0.89 \\
Traffic signals       & 0.96 & 0.92 \\
Gaze target           & 0.95 & 0.98 \\
Short-term intent     & 0.83 & 0.84 \\
\midrule
Structural compliance & \multicolumn{2}{c}{$96\%$ (after up to 2 prompt retries)} \\
\bottomrule
\end{tabular}
\end{table}

\paragraph{Failure-mode audit.}
We additionally conducted a 50-frame failure-mode audit on annotations that disagreed with the human reference.
The dominant error modes are missed pedestrians ($4\%$), misread traffic signals ($3\%$), and gaze-target mismatch ($2\%$); collectively the residual error rate stays below $10\%$, and the failures cluster on visually ambiguous scenes (distant pedestrians, partially occluded traffic lights, gaze near object boundaries).

\section{Multimodal Representation and Fusion}
\label{sec:appendix-modalities}

This section describes how the multimodal inputs used in our experiments are constructed from the EgoTraj dataset and how they are fused for trajectory forecasting.

\noindent\textbf{Ego-motion trajectory.}
The primary signal used for prediction was the past trajectory of the camera wearer, obtained from the headset pose measurements. At each timestep $t$, the pose consists of the 3D position $p_t \in \mathbb{R}^3$ and orientation $o_t \in \mathbb{R}^4$ represented as a quaternion.
Given an observation window of length $T_{obs}$, the past ego trajectory is defined as

\[
\mathcal{Y} = \{(p_t, o_t)\}_{t=t_0-T_{obs}+1}^{t_0}
\]

where $t_0$ denotes the final observation timestep.

\noindent\textbf{Social context.}
Human motion in crowded environments is influenced by nearby pedestrians. To model these interactions, we extracted trajectories of surrounding people from the egocentric RGB frames using pose detection and tracking. We examined three representations of nearby people: center points ($C$), bounding boxes ($B$), and full body poses ($P$). Among these, pose trajectories provided the most detailed motion cues for modeling human interactions.
For each detected pedestrian $n$, the trajectory is represented as

\[
J_n \in \mathbb{R}^{T_{obs} \times d}
\]

\noindent\textbf{Scene context.}
Environmental structure also constrains navigation behavior. To capture scene layout, we computed semantic segmentation maps from the RGB frames using a pretrained OneFormer model.
The segmentation mask at each timestep was embedded into a feature vector which encodes scene semantics such as walkable areas, obstacles, and building boundaries.

\[
S \in \mathbb{R}^{T_{obs} \times k}
\]

In addition to segmentation, we explored relative depth estimation derived from monocular depth prediction using Depth Anything V2. The resulting depth features are represented as

\[
D \in \mathbb{R}^{T_{obs} \times k}
\]

\noindent\textbf{Gaze representation.}
The MQPro provides binocular gaze direction estimates at each frame. To integrate gaze with visual observations, the 3D gaze vectors were projected into normalized image coordinates. This representation captured the visual attention of the camera wearer and provides cues about potential navigation targets.

\[
G = \{(u_t, v_t)\}_{t=t_0-T_{obs}+1}^{t_0}
\]

where $(u_t, v_t)$ denotes the gaze location in the image plane.

\noindent\textbf{Multimodal input formulation.}
The contextual information used by the model is defined as:

\[
Z = [Y, S, G],~~~\text{or alternatively}~~~Z = [Y, D, G],
\]

depending on whether semantic segmentation or depth features are used. Together with the ego trajectory $\mathcal{Y}$, these modalities formed the complete input for trajectory prediction.

\begin{figure}[t]
\centering
\includegraphics[width=0.9\textwidth]{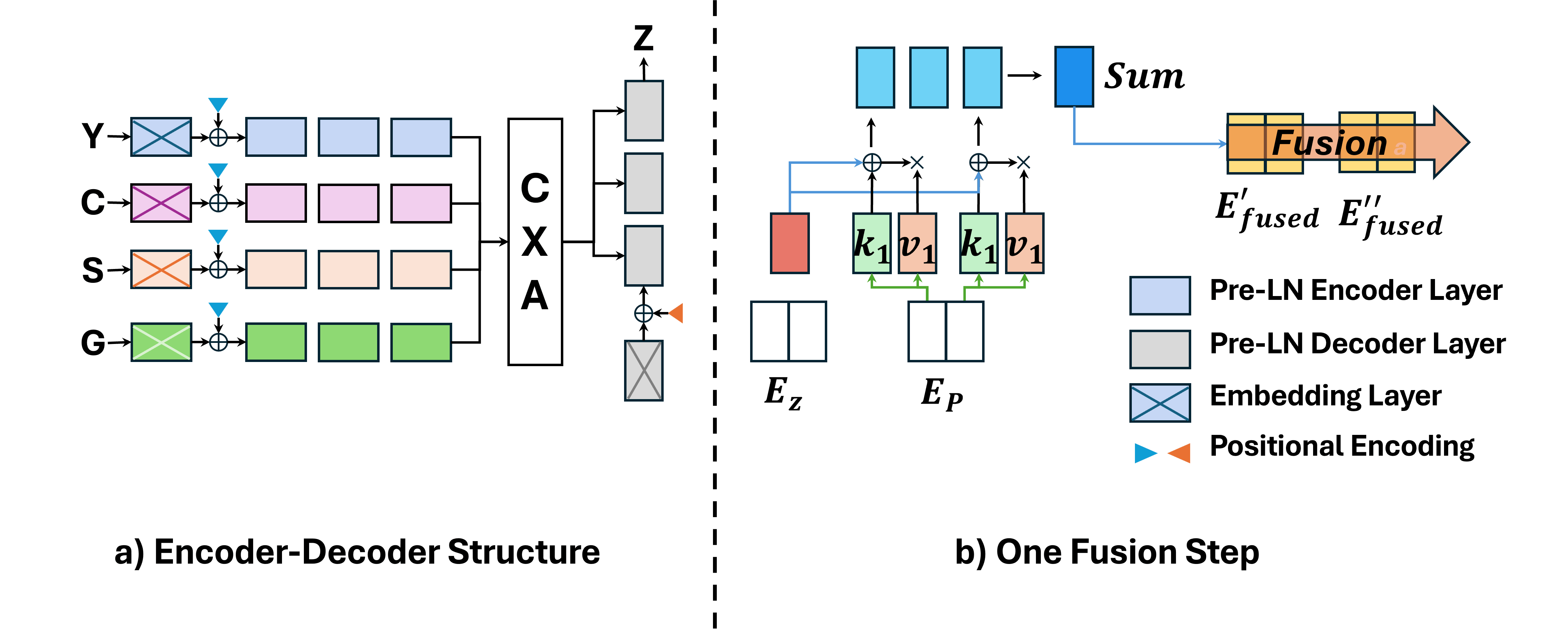}
\caption{(a) Overview of the proposed multimodal trajectory forecasting model. Each modality (ego-motion, social context, scene features, and gaze) is encoded by a dedicated transformer encoder stream. The encoded representations are fused and passed to a decoder that predicts the future trajectory of the camera wearer. (b) Illustration of the cascaded cross-attention mechanism used to progressively fuse the multimodal encodings.}
\label{fig:model_architecture}
\end{figure}

\noindent\textbf{Multimodal fusion.}
To integrate the information from different modalities, we employed a cascaded cross-attention fusion mechanism within a transformer-based architecture. In this design, the encoded representation of the ego trajectory acted as the query, while contextual modalities (nearby pedestrians, scene features, and gaze) served as keys and values. The cascaded structure sequentially incorporated information from each modality, allowing the model to dynamically attend to the most relevant cues. Figure~\ref{fig:model_architecture} shows the adopted model architecture with the multimodal cascaded cross-attention fusion mechanism.

Compared with simple concatenation-based fusion, cross-attention enabled the model to adaptively weight multimodal signals depending on the navigation context. Scene semantics provided structural constraints such as walkable regions and obstacles, social trajectories captured interactions with nearby pedestrians, and gaze signals reflected the camera wearer's visual attention and potential navigation intent. The effectiveness of this multimodal fusion strategy was reflected in the ablation study (Table~\ref{tab:ablation}), where progressively incorporating contextual modalities consistently improves prediction accuracy. In particular, both scene segmentation and gaze provided substantial individual improvements over ego-motion alone, while the full multimodal configuration ($\mathcal{Y}+P+S+G$) achieved the best overall performance.

\end{document}